\documentclass{article}

\PassOptionsToPackage{numbers, sort&compress}{natbib}

\usepackage[final]{neurips_2020}

\usepackage[utf8]{inputenc} %
\usepackage[T1]{fontenc}    %
\usepackage{hyperref}       %
\usepackage{url}            %
\usepackage{booktabs}       %
\usepackage{amsfonts}       %
\usepackage{nicefrac}       %
\usepackage{microtype}      %
\usepackage{graphicx}       %
\usepackage{subfig}         %
\usepackage{amsmath}        %
\usepackage{bm}             %
\usepackage{pgf}            %
\usepackage{tikz}           %
\usepackage{wrapfig}        %
\usepackage{multirow}       %
\usepackage{enumitem}       %
\usepackage{mathtools}      %

\usepackage{array}          %
\usepackage{float}          %

\newcommand{\adj}{\mathbf{A}}
\newcommand{\weight}{\mathbf{W}}
\newcommand{\features}{\mathbf{X}}
\newcommand{\featset}{\mathcal{X}}

\newcommand{\softout}{\mathbf{s}}
\newcommand{\neighbors}{\mathcal{N}}
\newcommand{\lone}{\text{L}_1}
\newcommand{\pertm}{\tilde{\mathbf{X}}_\epsilon}
\newcommand{\pertmset}{\tilde{\mathcal{X}}_\epsilon}

\newtheorem{theorem}{Theorem}

\newtheorem{corollary}{Corollary}
\newtheorem{lemma}{Lemma}
\newenvironment{proof}{}{$\square$}

\DeclareMathOperator*{\argmin}{arg\,min}

\newcommand{\sg}[1]{}

\graphicspath{ {./assets/} }

\title{Reliable Graph Neural Networks via\\ Robust Aggregation}

\author{%
  Simon~Geisler \qquad Daniel~Z\"ugner \qquad Stephan~G\"unnemann \\
  Department of Informatics\\
  Technical University of Munich\\
  \texttt{\{geisler, zuegnerd, guennemann\}@in.tum.de} \\
}

\begin{document}

\maketitle

\begin{abstract}
  Perturbations targeting the graph structure have proven to be extremely effective in reducing the performance of Graph Neural Networks (GNNs), and traditional defenses such as adversarial training do not seem to be able to improve robustness. This work is motivated by the observation that adversarially injected edges effectively can be viewed as additional samples to a node's neighborhood aggregation function, which results in distorted aggregations accumulating over the layers. Conventional GNN aggregation functions, such as a sum or mean, can be distorted arbitrarily by a single outlier. We propose a robust aggregation function motivated by the field of robust statistics. Our approach exhibits the largest possible breakdown point of 0.5, which means that the bias of the aggregation is bounded as long as the fraction of adversarial edges of a node is less than 50\%. Our novel aggregation function, Soft Medoid, is a fully differentiable generalization of the Medoid and therefore lends itself well for end-to-end deep learning. Equipping a GNN with our aggregation improves the robustness with respect to structure perturbations on Cora ML by a factor of 3 (and 5.5 on Citeseer) and by a factor of 8 for low-degree nodes.
\end{abstract}  

\section{Introduction}\label{sec:introduction}

Learning on graph data has gained strong attention in recent years, specifically powered by the success of graph neural networks~\cite{Kipf2017, Hamilton2017}. Like for classic neural networks, (non-)robustness to adversarial perturbations has shown to be a critical issue for GNNs as well~\citep{Dai2018, Zugner2018}. In contrast to other application domains of deep learning, adversaries on graphs are especially challenging because not only the attributes might be perturbed, but also the discrete structure. Recently, many effective attacks on graph neural networks have been proposed~\citep{Dai2018, Zugner2018, Xu2019a, Zugner2019a, Bojchevski2019, Waniek2018}, and there is strong evidence that attacking the graph structure is more effective than attacking the attributes \citep{Zugner2018, Wu2019}. 

While recent research suggests that effective defenses against attribute attacks can be found, e.g.\ robust training~\citep{Zugner2019a}, defenses against structure attacks remain an unsolved topic~\cite{Zuegner2020, Xu2019a, Dai2018}. Moreover, approaches such as~\citep{Wu2019, Entezari2020}, solely focus on defending against specific attack characteristics. On the contrary, \citet{Carlini2017a} show that heuristic defenses often can be bypassed. Thus, we design our model without attack-specific assumptions.

Message passing is the core operation powering modern GNNs~\citep{Gilmer2017}. In the message passing steps, a node's embedding is updated by aggregating over its neighbors' embeddings. In this regard, adversarially inserted edges add additional data points to the aggregation and therefore perturb the output of the message passing step. Standard aggregation functions like a sum can be arbitrarily distorted by only a single outlier. Thus, we reason that on top of the usual (potentially non-robust) neural network components, GNNs introduce additional (typically non-robust) aggregations. Note that many other countermeasures w.r.t.~adversarial vulnerability are orthogonal to our approach and can be applied additionally. 

We propose a novel robust aggregation function for GNNs to address this drawback. This aggregation function is novel in the context of deep learning. Our basic building block can be used within a large number of architectures by replacing the aggregation function with our proposed one.
Our robust location estimator Soft Medoid is smooth and differentiable, which makes it well-suited for being used within a neural network, and it has the best possible breakdown point of 0.5. With an appropriate budget, the adversary can only perturb a subset of the aggregation inputs with the goal of crossing the decision boundary. As long as the adversary only controls the minority of inputs, our robust estimator comes with a bounded error regardless of the attack characteristics (i.e.~no attack can distort the aggregation result arbitrarily). 

Empirically, our method improves the robustness of its base architecture w.r.t.\ structural perturbations by up to 550\% (relative), and outperforms previous state-of-the-art defenses. Moreover, we improve the robustness of the especially challenging to defend low degree nodes by a factor of 8.

\section{Robust aggregation functions for graph neural networks}\label{sec:preliminaries}

Throughout this work, we use the formulation in~\autoref{eq:mean-how-powerfull} (omitting the bias) for the message passing operation.
\begin{equation}\label{eq:mean-how-powerfull}
  \mathbf{h}^{(l)}_v = \sigma^{(l)} \left( \text{AGGREGATE}^{(l)} \left \{ \left( \adj_{vu}, \mathbf{h}^{(l-1)}_u \weight^{(l)} \right), \forall \, u\in \neighbors(v) \cup v \right \} \right)
\end{equation}
\(\mathbf{h}^{(l)}_v\) denotes the embedding of node \(v\) in the \(l\)-th layer; \(\mathbf{h}^{(0)}_v\) represents the (normalized) input features of node \(v\). Further, \( \adj \) is the (potentially normalized) message passing matrix, \( \weight^{(l)} \) the parameter for the trainable linear transformation and \(\sigma^{(l)}(\mathbf{z})\) the (non-linear) activation. \(\neighbors(v)\) is the set of neighbors of node \(v\). GCN \cite{Kipf2017} instantiates~\autoref{eq:mean-how-powerfull} as \(\mathbf{h}^{(l)}_v = \text{ReLU} ( \text{SUM} \{ (\adj_{vu} \mathbf{h}^{(l-1)}_u \weight^{(l)} ), \forall \, u\in \neighbors(v) \cup v \} ) \),  where \(\tilde{\mathbf{A}} = \hat{\mathbf{A}} + \mathbf{I}_N\), \(\hat{\mathbf{D}}_{ii}=\sum_j \tilde{\mathbf{A}}_{ij}\) and \(\adj = \hat{\mathbf{D}}^{\nicefrac{1}{2}}\tilde{\mathbf{A}}\hat{\mathbf{D}}^{\nicefrac{1}{2}}\) represents the normalization of the original adjacency matrix \(\hat{\mathbf{A}}\). Common examples for \(\text{AGGREGATE}^{(l)}\) are weighted mean\footnote{Technically we should call this operation weighted sum since the weights often do not sum up to 1. However, mean seems to be the widely accepted term (e.g.~see~\cite{Xu2019}).}~\citep{Kipf2017, Gao2019a, Velickovic2018, AbuElHaija2019}, the \(\max\) operation~\cite{Hamilton2017} or summation~\cite{Xu2019}. From a robust statistics point of view, a single perturbed embedding in \(v\)'s neighborhood suffices to arbitrarily deviate the resulting embedding \(\mathbf{h}^{(l)}_v\). We hypothesize that the non-robustness of the aggregation function contributes to GNNs' non-robustness.

\begin{figure}
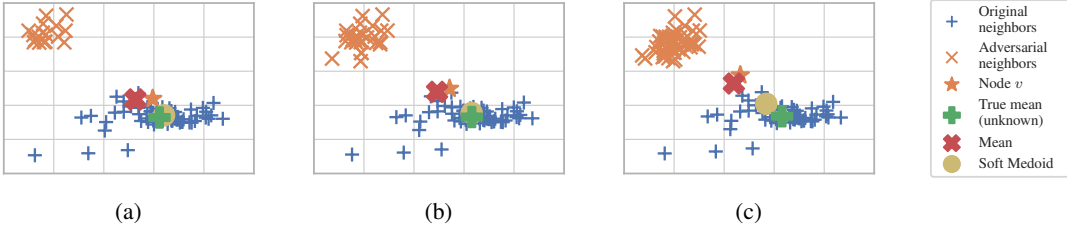

  \centering
  \centering
  \makebox[\columnwidth][c]{
    \(\begin{array}{cccc}
      \subfloat[]{\resizebox{0.27\linewidth}{!}{\input{assets/outlier_embedding_pert_no_legend17.pgf}}} &  
      \subfloat[]{\resizebox{0.27\linewidth}{!}{\input{assets/outlier_embedding_pert_no_legend29.pgf}}} & 
      \subfloat[]{\resizebox{0.27\linewidth}{!}{\input{assets/outlier_embedding_pert_no_legend50.pgf}}} &
      \resizebox{0.1575\linewidth}{!}{\input{assets/outlier_embedding_pert_legend.pgf}}\\
    \end{array}\)
  }
  \caption{ 
  We show the output layer (\(l=2\)) message passing step, i.e.\ the input of \(\text{AGGREGATE}^{(l)}\), for adversarially added edges of an exemplary node \(v\). The adversarial edges are obtained with a Nettack~\citep{Zugner2018} evasion attack (at test time).
  For a two-dimensional visualization we used PCA on the weighted node embeddings \(\adj_{sw} \mathbf{h}^{(l-1)}_w \weight^{(l)}\) of all edges \((s,w) \in \adj\), but solely plot \(v\)'s neighborhood. We show the aggregation for 17, 29, and 50 perturbations in figure (a) to (c), respectively.\label{fig:exemplarypredictions}}
\end{figure}

To back this hypothesis, we analyze the distortion of the neighborhood aggregation based on an exemplary message passing aggregation step in~\autoref{fig:exemplarypredictions}. The adversary inserts edges that result in a concentrated set of outliers. Only about 25\% of outliers in the aggregation suffice to move the output outside of the convex hull of the clean data points. We see that a robust location estimator, such as the proposed Soft Medoid in~\autoref{eq:softmaxweighteddist}, is much less affected by the outliers. Thus, we propose to use a \emph{robust} aggregation function in the message passing operation~\autoref{eq:mean-how-powerfull}. 

Robustness of a location estimator has multiple facets. The breakdown point \(\epsilon^*(t, \features)\) (see~\autoref{eq:breakdown})~\cite{Donoho1983} measures the percentage of perturbed data points \(\epsilon\) until the estimator \(t\) can be arbitrarily distorted. It is well studied and has a probabilistic motivation for algebraically tailed distributions~\citep{Lopuhaa1991}. Complementary, the maxbias curve \(B(\epsilon)\) (see~\autoref{eq:maxbias}) reports the maximum possible deviation of the location estimate between the clean and perturbed data w.r.t.\ the ratio of perturbed data~\cite{Croux2002}. Naturally, we desire a robust estimator to have a high breakdown point and low maxbias curve.

Measures such as the breakdown point are widely used as a proxy for the robustness of an estimator. While they analyze unbounded attacks, adversarially added edges in graph neural networks are, of course, not unbounded. However, for a strong/sufficient perturbation of the output, the attacker will likely perturb a neighborhood with nodes that have very different attributes/embeddings. Note that the magnitude of a structure perturbation is typically measured by the number of added or deleted edges (i.e. neighbors in~\autoref{eq:mean-how-powerfull}). We investigate unbounded perturbations as a worst-case analysis and bounded attacks in our empirical evaluation. As we are going to see in~\autoref{fig:empbiascurve}, a robust estimator typically comes with a lower error for bounded perturbations as well.

Many such robust location estimators are computationally expensive or hard to implement in a vectorized fashion, and not continuously differentiable~\cite{Lopuhaa1991, Tukey1960, Huber1964, Maronna1976, Davies1987, Rousseeuw1984, Donoho1992, Diakonikolas2019b}. In our experimentation, we found the M(R)CD estimator~\cite{Boudt2020} and a differentiable dimension-wise median implementation (based on soft sorting~\cite{Cuturi2019}) computationally too demanding for the repeated message passing operation. Moreover, estimators for high dimensions~\cite{Diakonikolas2017} did not filter many adversarially added edges (perhaps the number of inputs to an aggregation in a GNN is too low).

We conclude that existing robust location estimators are ill-suited for use within a neural network, as fast computation and differentiability are crucial. Therefore we propose a novel robust and fully differentiable location estimator and base our aggregation function on the Medoid \( t_{\text{Medoid}}(\features) = \argmin_{\mathbf{y} \in \featset} \sum_{j=1}^n \|\mathbf{x}_j - \mathbf{y}\|\), a multivariate generalization of the Median. In contrast to the \( \lone \)-Estimator \(t_{\lone}(\features) = \argmin_{\mathbf{y} \in \mathbb{R}^d} \sum_{j=1}^n \|\mathbf{x}_j - \mathbf{y}\|\), the Medoid constrains the domain of optimization from \(\mathbf{y} \in \mathbb{R}^d\) to the input data points (\( \mathbf{y} \in \featset \)). Throughout the paper, we denote the data matrix as \(\mathbf{X}\) and its set representation with \(\mathcal{X}\) interchangeably.

We propose a differentiable generalization of the Medoid replacing \( \argmin \) with a softmax to form a weighted average. That is,
\begin{equation}\label{eq:softmedoid}
t_{\text{Medoid}}(\features) = \argmin_{\mathbf{y} \in \features} \sum\nolimits_{j=1}^n \|\mathbf{x}_j - \mathbf{y}\| \approx \sum\nolimits_{i=1}^n \hat{\softout}_i \mathbf{x}_i = \hat{\softout}^\top\features =: t_{\text{SM}}(\features)\,.  
\end{equation}
The weights \(0 \leq \hat{\softout}_i \leq 1, \sum_i \hat{\softout}_i = 1\) are obtained via softmax of the data points' distances:
\begin{equation}\label{eq:softmaxdist}
    \hat{\softout}_i 
    = \frac{\exp{\left(-\frac{1}{T}\sum_{j=1}^n\|\mathbf{x}_j - \mathbf{x}_i\| \right )}}{\sum_{q=1}^n \exp{\left (-\frac{1}{T} \sum_{j=1}^n \|\mathbf{x}_j - \mathbf{x}_q\| \right )}}\,,
\end{equation}
where \(T\) is a temperature parameter controlling the steepness of the \(\argmin\) approximation. In this approximation, a point that has small distances to all other data points (i.e., a central data point) will have a large weight \(\hat{\softout}_i\), whereas remote points will have weights close to zero. For \(T \to 0 \) we recover the exact Medoid and for \(T \to \infty \) the sample mean. Further, the range of the Soft Medoid is no longer limited to the data points themselves; it is now limited to the real numbers enclosed by the convex hull of the data points, i.e. \(t_{\text{SM}}(\features) \in \mathcal{H}(\features)\). Furthermore, due to the Euclidean distance, the (Soft) Medoid is \emph{orthogonal equivariant} \(t_{\text{SM}}(\mathbf{Q}\features + \mathbf{v}) = \mathbf{Q}\,t_{\text{SM}}(\features) + \mathbf{v}\), with the orthogonal matrix \(\mathbf{Q}\) and the translation vector \(\mathbf{v} \in \mathbb{R}^d\).

\section{Robustness analysis}\label{sec:breakdown}

The (non-robust) sample mean and maximally robust Medoid are special cases of our smooth generalization of the Medoid (see~\autoref{eq:softmedoid}), depending on the choice of the softmax temperature \(T\). Naturally, this raises the question to what extent the Soft Medoid shares the robustness properties with the Medoid (or the non-robustness properties of the sample mean). %
In this section we show the non-obvious fact that regardless of the choice of \(T \in [0, \infty) \) the Soft Medoid has an asymptotic breakdown point of \(\epsilon^*(t_{\text{SM}}, \features) = 0.5\). As a corollary, the Soft Medoid comes with a guarantee on the embedding space. %
We conclude with a discussion of the influence of the temperature \(T\). w.r.t.\ the maxbias curve.

The (finite-sample) breakdown point states the minimal fraction \(\epsilon = \nicefrac{m}{n}\) with \(m\) perturbed examples, so that the result of the location estimator \(t(\features)\) can be arbitrarily placed~\citep{Donoho1983}:
\begin{equation}\label{eq:breakdown}
  \epsilon^*(t, \features) = \min_{1 \le m \le n} \left \{ \frac{m}{n}: \sup_{\pertm} \|t(\features)-t(\pertm)\| = \infty \right \}
\end{equation}

For this purpose, \( \pertm \) denotes the perturbed data. To obtain \( \pertm \) (equivalently \( \pertmset \)) we may select and change up to \(m\) (or an \(\epsilon\) fraction of) data points of \(\mathbf{x}_i \in \featset \) and leave the rest as they are. \citet{Lopuhaa1991} show that for affine/orthogonal equivariant estimators such as the \( \lone \)-Estimator, the best possible breakdown point is \(\epsilon^*(t_{\lone}, \features) = 0.5\). The sample mean, on the other side of the spectrum, has an asymptotic breakdown point of \(\epsilon^*(t_{\mu}, \features) = 0\). A single perturbed sample is sufficient to introduce arbitrary deviations from the sample mean`s true location estimate \(t_{\mu}(\features)\). 

\begin{theorem}\label{theorem:softmedoidbreakdown}
  Let \(\featset = \{ \mathbf{\mathbf{x}}_1, \dots, \mathbf{\mathbf{x}}_n\} \) be a collection of points in \(\mathbb{R}^d\) with finite coordinates and temperature \(T \in [0, \infty) \). Then the Soft Medoid location estimator (\autoref{eq:softmedoid}) has the finite sample breakdown point of \(\epsilon^*(t_{\text{SM}}, \features) = \nicefrac{1}{n} \lfloor \nicefrac{(n+1)}{2}\rfloor \) (asymptotically \( \lim_{n \to \infty} \epsilon^*(t_{\text{SM}}, \features) = 0.5 \)).
\end{theorem}

Our analysis addresses the somewhat general question: How well do we need to approximate the Medoid or \( \lone \)-Estimator to maintain its robustness guarantees? Despite many approximate algorithms exits~\citep{Cohen2016a, Chandrasekaran1989, Newling2017, Feldman2011, Chin2011, IndykPiotrStanfordUniversity2000, Parrilo2001, Badoiu2002, Har-Peled2007}, we are the first to address this problem:

\begin{lemma}\label{lemma:conditionapproxl1breakdown}
  Let \(\featset = \{ \mathbf{\mathbf{x}}_1, \dots, \mathbf{\mathbf{x}}_n\} \) be a collection of points in \(\mathbb{R}^d\), which are (w.l.o.g.) centered such that \ \(\hat{t}(\features)=0\). Then, the (orthogonal equivariant) approximate Medoid or \( \lone \)-Estimator  \( \hat{t} \) has a breakdown point of \(\epsilon^*(\hat{t}, \features) = \nicefrac{1}{n} \lfloor \nicefrac{(n+1)}{2}\rfloor \), if the following condition holds: \( \lim_{p \to \infty} \nicefrac{\hat{t}(\pertm)}{p} = \mathbf{0} \). Where
  \( \pertmset = \{ \tilde{\mathbf{x}}_1, \dots, \tilde{\mathbf{x}}_m,\,\mathbf{x}_{m+1}, \dots,\mathbf{x}_{n} \} \) is obtained from \(\featset\) by replacing \(m = \lfloor \nicefrac{(n-1)}{2} \rfloor \) \emph{arbitrary} samples with a point mass on the first axis: \(\tilde{\mathbf{x}}_i = [\begin{matrix}p & 0 & \cdots & 0\end{matrix}]^\top, \forall i \in \{1, \dots, m\}\).
\end{lemma}

\begin{wrapfigure}[15]{r}{0.5\textwidth}
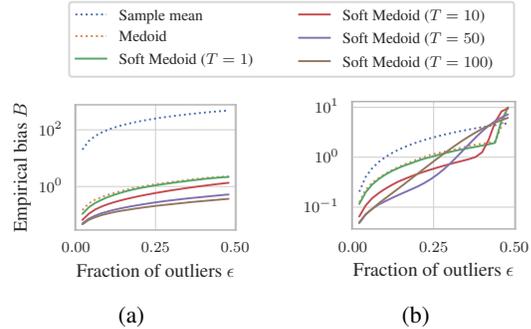

  \centering
  \vspace{-22pt}
  \hbox{\hspace{15pt} \resizebox{0.9\linewidth}{!}{
\begingroup%
\makeatletter%
\begin{pgfpicture}%
\pgfpathrectangle{\pgfpointorigin}{\pgfqpoint{3.375526in}{0.733252in}}%
\pgfusepath{use as bounding box, clip}%
\begin{pgfscope}%
\pgfsetbuttcap%
\pgfsetmiterjoin%
\definecolor{currentfill}{rgb}{1.000000,1.000000,1.000000}%
\pgfsetfillcolor{currentfill}%
\pgfsetlinewidth{0.000000pt}%
\definecolor{currentstroke}{rgb}{1.000000,1.000000,1.000000}%
\pgfsetstrokecolor{currentstroke}%
\pgfsetdash{}{0pt}%
\pgfpathmoveto{\pgfqpoint{0.000000in}{0.000000in}}%
\pgfpathlineto{\pgfqpoint{3.375526in}{0.000000in}}%
\pgfpathlineto{\pgfqpoint{3.375526in}{0.733252in}}%
\pgfpathlineto{\pgfqpoint{0.000000in}{0.733252in}}%
\pgfpathclose%
\pgfusepath{fill}%
\end{pgfscope}%
\begin{pgfscope}%
\pgfsetbuttcap%
\pgfsetmiterjoin%
\definecolor{currentfill}{rgb}{1.000000,1.000000,1.000000}%
\pgfsetfillcolor{currentfill}%
\pgfsetfillopacity{0.800000}%
\pgfsetlinewidth{1.003750pt}%
\definecolor{currentstroke}{rgb}{0.800000,0.800000,0.800000}%
\pgfsetstrokecolor{currentstroke}%
\pgfsetstrokeopacity{0.800000}%
\pgfsetdash{}{0pt}%
\pgfpathmoveto{\pgfqpoint{0.122222in}{0.100000in}}%
\pgfpathlineto{\pgfqpoint{3.253304in}{0.100000in}}%
\pgfpathquadraticcurveto{\pgfqpoint{3.275526in}{0.100000in}}{\pgfqpoint{3.275526in}{0.122222in}}%
\pgfpathlineto{\pgfqpoint{3.275526in}{0.611030in}}%
\pgfpathquadraticcurveto{\pgfqpoint{3.275526in}{0.633252in}}{\pgfqpoint{3.253304in}{0.633252in}}%
\pgfpathlineto{\pgfqpoint{0.122222in}{0.633252in}}%
\pgfpathquadraticcurveto{\pgfqpoint{0.100000in}{0.633252in}}{\pgfqpoint{0.100000in}{0.611030in}}%
\pgfpathlineto{\pgfqpoint{0.100000in}{0.122222in}}%
\pgfpathquadraticcurveto{\pgfqpoint{0.100000in}{0.100000in}}{\pgfqpoint{0.122222in}{0.100000in}}%
\pgfpathclose%
\pgfusepath{stroke,fill}%
\end{pgfscope}%
\begin{pgfscope}%
\pgfsetbuttcap%
\pgfsetroundjoin%
\pgfsetlinewidth{1.003750pt}%
\definecolor{currentstroke}{rgb}{0.298039,0.447059,0.690196}%
\pgfsetstrokecolor{currentstroke}%
\pgfsetdash{{1.000000pt}{1.650000pt}}{0.000000pt}%
\pgfpathmoveto{\pgfqpoint{0.144444in}{0.544383in}}%
\pgfpathlineto{\pgfqpoint{0.366667in}{0.544383in}}%
\pgfusepath{stroke}%
\end{pgfscope}%
\begin{pgfscope}%
\definecolor{textcolor}{rgb}{0.150000,0.150000,0.150000}%
\pgfsetstrokecolor{textcolor}%
\pgfsetfillcolor{textcolor}%
\pgftext[x=0.455556in,y=0.505495in,left,base]{\color{textcolor}\rmfamily\fontsize{8.000000}{9.600000}\selectfont Sample mean}%
\end{pgfscope}%
\begin{pgfscope}%
\pgfsetbuttcap%
\pgfsetroundjoin%
\pgfsetlinewidth{1.003750pt}%
\definecolor{currentstroke}{rgb}{0.866667,0.517647,0.321569}%
\pgfsetstrokecolor{currentstroke}%
\pgfsetdash{{1.000000pt}{1.650000pt}}{0.000000pt}%
\pgfpathmoveto{\pgfqpoint{0.144444in}{0.389450in}}%
\pgfpathlineto{\pgfqpoint{0.366667in}{0.389450in}}%
\pgfusepath{stroke}%
\end{pgfscope}%
\begin{pgfscope}%
\definecolor{textcolor}{rgb}{0.150000,0.150000,0.150000}%
\pgfsetstrokecolor{textcolor}%
\pgfsetfillcolor{textcolor}%
\pgftext[x=0.455556in,y=0.350562in,left,base]{\color{textcolor}\rmfamily\fontsize{8.000000}{9.600000}\selectfont Medoid}%
\end{pgfscope}%
\begin{pgfscope}%
\pgfsetroundcap%
\pgfsetroundjoin%
\pgfsetlinewidth{1.003750pt}%
\definecolor{currentstroke}{rgb}{0.333333,0.658824,0.407843}%
\pgfsetstrokecolor{currentstroke}%
\pgfsetdash{}{0pt}%
\pgfpathmoveto{\pgfqpoint{0.144444in}{0.228982in}}%
\pgfpathlineto{\pgfqpoint{0.366667in}{0.228982in}}%
\pgfusepath{stroke}%
\end{pgfscope}%
\begin{pgfscope}%
\definecolor{textcolor}{rgb}{0.150000,0.150000,0.150000}%
\pgfsetstrokecolor{textcolor}%
\pgfsetfillcolor{textcolor}%
\pgftext[x=0.455556in,y=0.190093in,left,base]{\color{textcolor}\rmfamily\fontsize{8.000000}{9.600000}\selectfont Soft Medoid (\(\displaystyle T=1\))}%
\end{pgfscope}%
\begin{pgfscope}%
\pgfsetroundcap%
\pgfsetroundjoin%
\pgfsetlinewidth{1.003750pt}%
\definecolor{currentstroke}{rgb}{0.768627,0.305882,0.321569}%
\pgfsetstrokecolor{currentstroke}%
\pgfsetdash{}{0pt}%
\pgfpathmoveto{\pgfqpoint{1.739846in}{0.544383in}}%
\pgfpathlineto{\pgfqpoint{1.962068in}{0.544383in}}%
\pgfusepath{stroke}%
\end{pgfscope}%
\begin{pgfscope}%
\definecolor{textcolor}{rgb}{0.150000,0.150000,0.150000}%
\pgfsetstrokecolor{textcolor}%
\pgfsetfillcolor{textcolor}%
\pgftext[x=2.050957in,y=0.505495in,left,base]{\color{textcolor}\rmfamily\fontsize{8.000000}{9.600000}\selectfont Soft Medoid (\(\displaystyle T=10\))}%
\end{pgfscope}%
\begin{pgfscope}%
\pgfsetroundcap%
\pgfsetroundjoin%
\pgfsetlinewidth{1.003750pt}%
\definecolor{currentstroke}{rgb}{0.505882,0.447059,0.701961}%
\pgfsetstrokecolor{currentstroke}%
\pgfsetdash{}{0pt}%
\pgfpathmoveto{\pgfqpoint{1.739846in}{0.377744in}}%
\pgfpathlineto{\pgfqpoint{1.962068in}{0.377744in}}%
\pgfusepath{stroke}%
\end{pgfscope}%
\begin{pgfscope}%
\definecolor{textcolor}{rgb}{0.150000,0.150000,0.150000}%
\pgfsetstrokecolor{textcolor}%
\pgfsetfillcolor{textcolor}%
\pgftext[x=2.050957in,y=0.338855in,left,base]{\color{textcolor}\rmfamily\fontsize{8.000000}{9.600000}\selectfont Soft Medoid (\(\displaystyle T=50\))}%
\end{pgfscope}%
\begin{pgfscope}%
\pgfsetroundcap%
\pgfsetroundjoin%
\pgfsetlinewidth{1.003750pt}%
\definecolor{currentstroke}{rgb}{0.576471,0.470588,0.376471}%
\pgfsetstrokecolor{currentstroke}%
\pgfsetdash{}{0pt}%
\pgfpathmoveto{\pgfqpoint{1.739846in}{0.211104in}}%
\pgfpathlineto{\pgfqpoint{1.962068in}{0.211104in}}%
\pgfusepath{stroke}%
\end{pgfscope}%
\begin{pgfscope}%
\definecolor{textcolor}{rgb}{0.150000,0.150000,0.150000}%
\pgfsetstrokecolor{textcolor}%
\pgfsetfillcolor{textcolor}%
\pgftext[x=2.050957in,y=0.172215in,left,base]{\color{textcolor}\rmfamily\fontsize{8.000000}{9.600000}\selectfont Soft Medoid (\(\displaystyle T=100\))}%
\end{pgfscope}%
\end{pgfpicture}%
\makeatother%
\endgroup%
}}
  \vspace{-14pt}
  \makebox[\linewidth][c]{
  \(\begin{array}{cc}
    \subfloat[]{\resizebox{0.50\linewidth}{!}{\input{assets/emp_bias_curve_no_legend_1e3.pgf}}} & 
    \subfloat[]{\resizebox{0.4825\linewidth}{!}{\input{assets/emp_bias_curve_no_leglab_1e1.pgf}}} \\
  \end{array}\)
  }
  \caption{Empirical bias \(B(\epsilon)\), for 50 samples from a centered (\(t_{\text{SM}}(\features) = 0\)) bivariate normal distribution. (a) shows the bias for a perturbation with norm 1000, and (b) 10.\label{fig:empbiascurve}}
\end{wrapfigure}

As a direct consequence of \autoref{lemma:conditionapproxl1breakdown}, it is not decisive how closely we approximate the true Medoid. The condition rather imposes an upper bound on the growth of the location estimator over the magnitude of the perturbation \(p\). In addition to the formal proof in~\autoref{sec:appendix_proof_breakdown_sm}, we now present an illustrative proof sketch for a simplified scenario, which highlights why the Soft Medoid has such a strong guarantee regardless of \(T \in [0, \infty) \) and omits the detour via~\autoref{lemma:conditionapproxl1breakdown}.

\begin{proof}\textit{Proof Sketch}\label{proof:sketch}
  Due to the orthogonal equivariance we may choose \(t_{\text{SM}}(\features) = 0\), without loss of generality. Let \( \pertmset \) be decomposable such that \(\pertmset = \pertmset^{(\text{clean})} \cup \pertmset^{(\text{pert.})} \). Clearly the worst-case perturbation is obtained when \(\pertmset^{(\text{pert.})}\) concentrates on a point mass~\cite{Croux2002}. Due to orthogonal equivariance we can, thus, pick \(\tilde{\mathbf{x}}_i = [\begin{matrix}p & 0 & \cdots & 0\end{matrix}]^\top, \forall \tilde{\mathbf{x}}_i \in \pertmset^{(\text{pert.})}\) w.l.o.g.
  In the following, we analyze the special case where all clean data points are located in the origin \(\mathbf{x}_{i} = \mathbf{0}, \forall \mathbf{x}_{i} \in \pertmset^{(\text{clean})}\). 
  
  We now have to find the minimal fraction of outliers \(\epsilon\) for which \(\lim_{p \to \infty} \|t_{\text{SM}}(\pertm)\| < \infty\) does not hold anymore. 
  Here, both terms in the equation of the Soft Medoid \(t_{\text{SM}}(\pertm) = \hat{\softout}^\top\pertm\) depend on \(p\) and \(\lim_{p \to \infty} t_{\text{SM}}(\pertm) = \lim_{p \to \infty} \hat{\softout}^\top\pertm\) leads to the undefined case of \(0 \cdot \infty\).
  However, because of \(\lim_{x \to \infty} x e^{-x/a} = 0\) for \(a \in [0, \infty)\), it turns out that we just have to analyze \(\hat{\softout}\) for \(p \to \infty\). That is, if \(\hat{\softout}^{(\text{pert.})} \to 0\) the perturbed data have zero weight in the aggregation. We now relate the weight of any perturbed data point \(\softout^{(\text{pert.})}\) to the weight of any clean data point \(\hat{\softout}^{(\text{clean})}\):
  \[
      \frac{\hat{\softout}^{(\text{pert.})}}{\hat{\softout}^{(\text{clean})}} 
      = \frac{\exp \left \{-\frac{1}{T} \sum_{\tilde{\mathbf{x}}_j \in \pertmset} \|\tilde{\mathbf{x}}_j - \tilde{\mathbf{x}}^{(\text{pert.})}\| \right \}}{\exp \left \{-\frac{1}{T} \sum_{\tilde{\mathbf{x}}_j \in \pertmset} \|\tilde{\mathbf{x}}_j - \tilde{\mathbf{x}}^{(\text{clean})}\|\right \}} 
      = \exp \Biggl\{-\frac{1}{T} \underbrace{\Biggl[ \Biggl(\, \sum_{\tilde{\mathbf{x}}_j \in \pertmset^{(\text{clean})}} p \Biggr) - \Biggl(\, \sum_{\tilde{\mathbf{x}}_j \in \pertmset^{(\text{pert.})}} p \Biggr) \Biggr]}_{\left( |\pertmset^{(\text{clean})}| - |\pertmset^{(\text{pert.})}|\right) \cdot p } \Biggr\}
  \]
  If we have more clean points than perturbed points (\(|\pertmset^{(\text{clean})}| > |\pertmset^{(\text{pert.})}|\)), then \(\lim_{p \to \infty} \nicefrac{\hat{\softout}^{(\text{pert.})}}{\hat{\softout}^{(\text{clean})}} = \exp(-\infty) = 0\). Note that \(\nicefrac{\hat{\softout}^{(\text{pert.})}}{\hat{\softout}^{(\text{clean})}} = 0\) can only be true if \(\hat{\softout}^{(\text{pert.})} = 0\). Hence, the norm of the Soft Medoid is finite when the perturbation \(p\) approaches infinity iff \(\epsilon < 0.5\).
\end{proof}

For a corollary of~\autoref{theorem:softmedoidbreakdown}, we formally introduce the (asymptotic) maxbias curve
\begin{equation}\label{eq:maxbias}
  B^*(\epsilon, t, \mathcal{D}_\featset) = \sup_{H} \|t(\mathcal{D}_\featset) - t\left((1 - \epsilon)\mathcal{D}_\featset + \epsilon H \right)\|\,,
\end{equation}
with the data distribution \(\mathcal{D}_\featset\) and arbitrary distribution \(H\) representing the perturbation. The maxbias curve models the maximum deviation between clean and perturbed estimate over different percentages of perturbations \(\epsilon\). From~\autoref{theorem:softmedoidbreakdown} and the monotonicity of the maxbias curve, \autoref{corollary:finitemaxbias} follows.
\begin{corollary}\label{corollary:finitemaxbias}
  Let \(\featset = \{ \mathbf{\mathbf{x}_1}, \dots, \mathbf{\mathbf{x}_n}\} \) be a collection of points in \(\mathbb{R}^d\) with finite coordinates and the constant temperature \(T \in [0, \infty) \). Then the Soft Medoid location estimator (\autoref{eq:softmedoid}) has a finite maxbias curve \( B^*(\epsilon, t_{\text{SM}}, \mathcal{D}_\featset) < \infty \) for \(\epsilon < \epsilon^*(t_{\text{SM}}, \features)\).
\end{corollary}
There exists a finite upper bound on the maxbias, i.e.\ the maximum deviation \(\|t_{\text{SM}}(\features) - t_{\text{SM}}(\pertm)\| < \infty\) between the estimate on the clean data \(\features\) and perturbed data \(\pertm\) is limited. Consequently, using the Soft-Medoid translates to robustness guarantees on the embedding space of each layer. However, deriving this upper bound analytically is out of scope for this work.

In \autoref{fig:empbiascurve}, we give empirical results for a fixed point mass perturbation on the first axis over increasing values of \( \epsilon \). \autoref{fig:empbiascurve} (a) shows that for high temperatures and distant perturbations our Soft Medoid achieves an even lower bias than the Medoid because it essentially averages the clean points. (b) shows that this comes with the risk of a higher bias for small perturbations and high \( \epsilon \). However, in case the perturbation is close to the data points, the bias cannot be very high. In conclusion, in the context of a GNN and for an appropriate choice of \(T\) as well as \textit{bounded perturbations}, the Soft Medoid can help mitigate the effects of adversarially injected edges as long as \(\epsilon\) is sufficiently small.

\section{Instantiating the Soft Medoid for graph neural networks}\label{sec:softmedoid}

Before we can show the effectiveness of our method, we need to discuss how we can use the proposed Soft Medoid in GNNs. Effectively, we have to extend ~\autoref{eq:softmedoid} to the weighted case due to the weights in the respective message passing matrix \( \adj \):

\vspace{-7pt}
\begin{minipage}{.4\linewidth}
\begin{equation}\label{eq:resulting-wsm}
    \tilde{t}_{\text{WSM}}(\features, \mathbf{a}) = c\,(\softout \circ \mathbf{a})^\top\features
\end{equation}
\end{minipage}
\begin{minipage}{.55\linewidth}
\begin{equation}\label{eq:softmaxweighteddist}
    \softout_i = \frac{\exp{\left(-\frac{1}{T}\sum_{j=1}^n \mathbf{a}_j\|\mathbf{x}_j - \mathbf{x}_i\| \right )}}{\sum_{q=1}^n \exp{\left (-\frac{1}{T} \sum_{j=1}^n  \mathbf{a}_j\|\mathbf{x}_j - \mathbf{x}_q\| \right )}}
\end{equation}
\end{minipage}%

\noindent where \(\mathbf{a}\) is a non-negative weight vector (e.g.\ the weights in a row of \( \adj \)) and %
\(c=\nicefrac{(\sum_{j=1}^n \mathbf{a}_j)}{(\sum_{j=1}^n \mathbf{s}_j\mathbf{a}_j)} \).

Since the Soft Medoid interpolates between the Medoid and mean, we indeed have to adapt the location estimator at two places: The generalized definition of \(\softout\) handles the weighted Medoid, while \(\softout \circ \mathbf{a}\) resembles the weighted mean (note that for \(T \to \infty \) all elements of \(\softout\) are equal, thus, using only \(\softout\) would result in an \emph{unweighted} mean; \(\softout \circ \mathbf{a}\) makes it a weighted mean). The multiplication with \(c\) simply ensures a proper normalization of \(\tilde{t}_{\text{WSM}}\) like in a standard GNN. %

\autoref{theorem:softmedoidbreakdown} holds for the weighted case accordingly: Given a weight vector \(\mathbf{a}\) with positive weights, the estimate \(\tilde{t}_{\text{WSM}}\) cannot be arbitrarily perturbed if \(\sum \mathbf{a}^{(\text{pert.})} < \sum \mathbf{a}^{(\text{clean})}\) is satisfied (see~\autoref{sec:appendix_proof_weightedsoftmedoid}). 

\begin{wraptable}[14]{r}{0.55\textwidth}
  \centering
  \caption{Average duration (time cost in ms) of one training epoch (over 200 epochs, preprocessing counts once). For the other defenses we used \href{https://github.com/DSE-MSU/DeepRobust}{DeepRobust's implementation}. We report ``-'' for an OOM. We used one 2.20 GHz core and one GeForce GTX 1080 Ti (11 Gb). For hyperparameters see~\autoref{sec:experiments}.\label{tab:runtime}}
  \resizebox{\linewidth}{!}{
      \begin{tabular}{lcccccc}
      \toprule
      {} & \multicolumn{2}{l}{Cora ML~[46]} & \multicolumn{2}{l}{Citeseer~[47]} & \multicolumn{2}{l}{PubMed~[46]} \\
      GDC Prepr. & & \checkmark &  & \checkmark & & \checkmark \\
      \midrule
      SM GCN      &                    41.2 &      210.9 &                          36.6 &      154.1 &                     86.0 &      497.8 \\
      SVD GCN     &                   119.4 &      120.8 &                          66.3 &       67.3 &                      - &          - \\
      Jaccard GCN &                    19.1 &      147.8 &                          11.2 &      118.0 &                   84.9 &      585.4 \\
      RGCN        &                     8.7 &        7.5 &                           6.3 &        9.3 &                      135.5 &         136.6  \\
      Vanilla GCN &                     5.1 &        7.1 &                           4.7 &        7.8 &                      6.0 &       66.1 \\
      Vanilla GAT &                    15.2 &       65.6 &                          11.8 &       53.3 &                   46.4 &      270.8 \\
      \bottomrule
      \end{tabular}
  }
\end{wraptable}

In \autoref{eq:mean-how-powerfull} we plug in the newly derived Weighted Soft Medoid \( \tilde{t}_{\text{WSM}}(\features, \mathbf{a}) \) for the \( \text{AGGREGATION} \). Thus, for node \(v\) in layer \(l\), \(\features\) represents the stacked embeddings \(\{\mathbf{h}^{(l-1)}_u \weight^{(l)}, \forall \, u\in \neighbors(v) \cup v\}\), and \(\mathbf{a}\) the weight vector consists of \(\{\adj_{vu}, \forall \, u\in \neighbors(v) \cup v\}\). Hence, we can think about the terms before \(\mathbf{X}\) in~\autoref{eq:resulting-wsm} as an input-dependent reweighting of the message passing matrix.

A sparse matrix implementation of the Weighted Soft Medoid has a time complexity of \(O(n \sum_{v=1}^n {(\text{deg}(v) + 1)}^2)\), with number of nodes \(n\). Due to the power law distribution of many graphs, we will also have a few nodes with a very large degree. To circumvent this issue and to enable a fully vectorized implementation we propose to calculate the Weighted Soft Medoid for the embeddings of the \(k\) neighbors with largest weight. This yields a time and space complexity of \(O(n k^2)\) and for \(k \ll n\) leads to a total worst-case complexity of \(O(n)\). The time cost of the Soft Medoid (SM GCN) is comparable to the defenses SVD GCN and Jaccard GCN (see~\autoref{tab:runtime}).

\section{Experimental evaluation}\label{sec:experiments}

In~\autoref{sec:influencetemperature}, we discuss the influence of the temperature \(T\).
While our main focus is on evaluating \emph{certifiable} robustness, we also analyze the empirical robustness via attacks (\autoref{sec:empirircalrobustness}). In~\autoref{sec:experimentalresults} we present the main results and comparison to other defenses. We mainly highlight results on Cora ML and attacks jointly adding and deleting edges (for other datasets/attacks see~\autoref{sec:appendix_detailed_results}). The source code is available at \url{https://www.daml.in.tum.de/reliable_gnn_via_robust_aggregation}.

\subsection{Setup}\label{sec:expsetup}

\textbf{Architectures.}
We compare our approach against the current state of the art defenses against structure attacks~\citep{Entezari2020, Wu2019, Zhu2019}. \emph{SVD GCN}~\citep{Entezari2020} performs a low-rank approximation of the adjacency matrix with a truncated SVD (the result is not sparse in general, we use rank 50), \emph{Jaccard GCN}~\citep{Wu2019} use the Jaccard similarity on the attributes to filter dissimilar edges (we use a similarity threshold of 0.01), and the \emph{RGCN}~\citep{Zhu2019} models the graph convolution via a Gaussian distribution for absorbing the effects of adversarial changes. Further, we compare the robustness to the general-purpose GNNs Graph Attention Network (GAT)~\citep{Velickovic2018}, %
Graph Diffusion Convolution (GDC) with a GCN architecture~\cite{Klicpera2019a}, and GCN~\citep{Kipf2017}. As baselines of robust location estimators, we equip a GCN and a GDC with the dimension-wise Median and the Medoid. Note that because of their non-differentiability, only the gradient for the selected/central item is non-zero---similarly to, e.g., max-pooling on images.

\textbf{Datasets.}
We evaluate these models on Cora ML~\citep{Sen2008}, Citeseer~\citep{McCallum2000}, and PubMed~\citep{Sen2008} for semi-supervised node classification. \autoref{sec:appendix_datasets} gives a summary of the size of the respective largest connected component, which we are using. None of the referenced attacks/defenses~\citep{Dai2018, Zugner2018, Xu2019a, Zugner2019a, Bojchevski2019, Waniek2018, Wu2019, Entezari2020, Miller2020} uses a larger dataset. Note that our approach scales (runtime/space) with \(\mathcal{O}(n)\) while SVD GCN has space complexity of \(\mathcal{O}(n^2)\).

\textbf{Hyperparameters.} We use two-layer GNNs with default parameters, as suggested by the respective authors for all the models. We use the personalized PageRank version of GDC. For a fair comparison, we set the number of hidden units for all architectures to 64, the learning rate to \(0.01\), weight decay to \(5\mathrm{e}{-4}\), and train for 3000 epochs with a patience of 300. For the architectures incorporating our Soft Medoid, we perform a grid search over different temperatures \(T\) (for the range of the temperatures \(T\) see~\autoref{fig:temperatureinfluence}). In case we are using GDC, we also test different values for the teleport probability \(\alpha \in [0.05, 0.4]\). In the experiments on Cora ML and Citeseer we use \(\alpha=0.15\) as well as \(k=64\). We use \(\alpha=0.15\) as well as \(k=32\) in the PubMed experiments. For each approach and dataset, we rerun the experiment with three different seeds, use each 20 labels per class for training and validation, and report the one-sigma error of the mean.

\textbf{Robustness certificates.} To measure certifiable robustness, we use Randomized Smoothing~\citep{Li2018, Lecuyer2019, Cohen2019} for GNNs~\citep{Bojchevski2020}. Randomized smoothing is a probabilistic, black-box robustness certification technique that is applicable to any model. Following~\citet{Bojchevski2020} we create an ensemble of models \(g(\mathbf{x})\) (aka the smooth classifier) that consists of the trained base classifier \(f(\mathbf{x})\) with random inputs. We randomly perturbed the input via independent random flips of elements in the binary feature matrix and/or adjacency matrix. For adding an element we use the probability \(p_a\) and for removing an element we use \(p_d\).

We treat the prediction of a smooth classifier as \emph{certifiably correct} if it is both correct and certifiably robust; i.e. the prediction does not change w.r.t.\ any of the considered perturbations/attacks. We refer to the ratio of certifiably correct predictions as the \emph{certification ratio} \(R(r_a, r_d)\) at addition radius \(r_a\) and deletion radius \(r_d\). For example, \(R(r_a=2,r_d=0)\) denotes the ratio of nodes that are robust (and correct) under insertion of any two edges. Higher is better. We compare the robustness for three different cases: (a) addition or deletion of edges, (b) only deletion, (c) only addition. For further details on randomized smoothing, we refer to~\autoref{sec:appendix_setupdetails}.

Comparing all these certification ratios \(R(r_a, r_d)\) at different radii is somewhat cumbersome and subjective. Therefore, we propose the accumulated certifications  
\begin{equation}\label{eq:accumcert}
  \text{AC} =  -R(0, 0) + \sum\nolimits_{r_a, r_d} R(r_a, r_d)
\end{equation}
as a single measure that captures overall robustness. We decide to subtract \(R(0, 0)\), because it reflects the accuracy of the smooth classifier. This metric is related to the area underneath the bivariate certification ratio \(R(r_a, r_d)\). Note that a more robust model has higher accumulated certifications.

To capture what certifiable radii one obtains for correct predictions, in~\autoref{tab:results}, we additionally report the average certifiable radii \(\bar{r}_a\) (and \(\bar{r}_d\)):
\begin{equation}\label{eq:certradius}
  \bar{r}_a := \frac{1}{|C|}\sum\nolimits_{i \in C} r_a^{\max}(i) \,.
\end{equation}
Here, \( C\) denotes the set of all correctly predicted nodes and \(r_a^{\max}(i)\) the maximum addition radius so that node $i$ can still be certified w.r.t.\ the smooth classifier \(g(\mathbf{x}_i)\); analogously for \(r_d\). The higher the better. 

\subsection{The temperature hyperparameter}\label{sec:influencetemperature}

\begin{wrapfigure}[13]{r}{.3625\textwidth}
  \centering
  \vspace{-36pt}
  \resizebox{\linewidth}{!}{\input{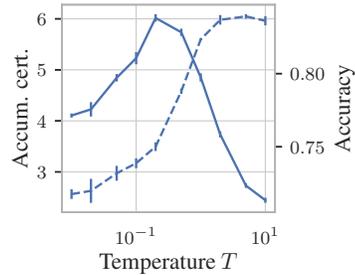}}
  \captionof{figure}{Influence of the temperature \(T\) on the accumulated certifications (solid) and accuracy of the base classifier (dashed).\label{fig:temperatureinfluence}}
\end{wrapfigure}

Following up on the concluding statement of~\autoref{sec:breakdown}, the temperature \(T\) is a central hyperparameter for a GNN equipped with a Soft Medoid. Our best-performing setup is a GDC equipped with a Soft Medoid (see~\autoref{sec:experimentalresults}). Consequently, we use this model for the analysis of the influence of \(T\).

\autoref{fig:temperatureinfluence} illustrates this relationship for a wide range of \(T\). Decreasing the temperature comes with increased robustness but at the cost of the accuracy. However, we cannot increase the robustness indefinitely and observe a maximum around \(T=0.2\). We hypothesize that this is because for too low values of \(T\) the Soft Medoid ignores all but one input and for high temperatures \(T\) we approach the non-robust sample mean.
In reference to~\autoref{sec:breakdown}, for the right temperature w.r.t.~the magnitude of perturbations, we essentially average over the clean data points. Depending on the requirements for the robustness accuracy trade-off, we conclude that the sweet spot is likely to be in the interval of \(T \in [0.2, 1]\).
With that in mind, we decide to report the reasonable trade-offs of \(T=1\), \(T=0.5\), and our most robust model  (\(T=0.2\)), for the experiments.

\subsection{Empirical robustness}\label{sec:empirircalrobustness}
\begin{figure}[ht]
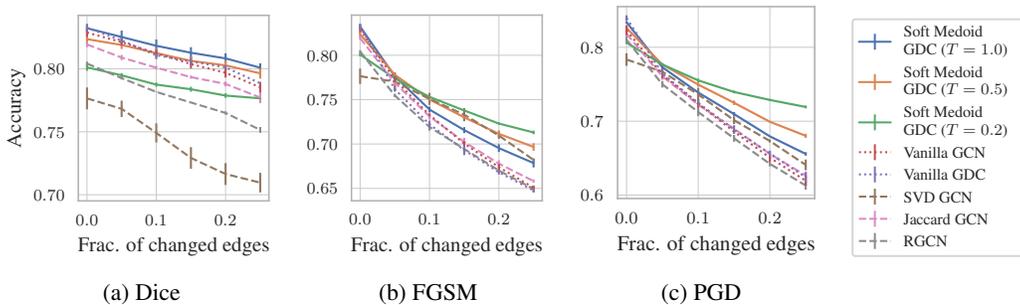

  \centering
  
  \makebox[\columnwidth][c]{
    \(\arraycolsep=0pt
    \begin{array}{cccc} 
    \subfloat[Dice]{\resizebox{0.285\linewidth}{!}{\input{assets/global_neurips_dice_cora_ml_acc_no_legend.pgf}}} &
    \subfloat[FGSM]{\resizebox{0.26\linewidth}{!}{\input{assets/global_neurips_fsgm_cora_ml_acc_no_leglab.pgf}}} &
    \subfloat[PGD]{\resizebox{0.26\linewidth}{!}{\input{assets/global_neurips_pgd_cora_ml_acc_no_leglab.pgf}}} &
    \resizebox{0.19\linewidth}{!}{\input{assets/global_neurips_fsgm_cora_ml_acc_legend.pgf}} \\
    \end{array}\)
  }
  
  \caption{Accuracy for evasion (transfer) attacks on Cora ML.\label{fig:globalfsgm}}
\end{figure}

The advantage of analyzing certifiable robustness is that it does not rely on specific attack approaches and the respective characteristic. However, the certificates we obtain are strictly speaking for the resulting smooth classifier. As~\citet{Cohen2019} point out, only a base classifier that is robust w.r.t.\ these small perturbations can result in a robust smooth classifier. 
Still, for completeness, we report in \autoref{fig:globalfsgm} the (empirical) robustness of the base classifier, i.e. we measure how the accuracy drops when attacking the adjacency matrix. In such a scenario one has to refer to a specific attack approach. As shown, our approach outperforms all baselines with a significant margin for strong perturbations. That is, the accuracy stays high despite many perturbed edges. We report the perturbed accuracy for Dice~\cite{Waniek2018}, a FGSM-like~\cite{Goodfellow2015} attack that greedily flips the element in \(\adj\) which contributes most to the test loss and Projected Gradient Descent (PGD) for \(L_0\) perturbations~\cite{Xu2019a}. For Nettack~\cite{Zugner2018}, Metattack~\citep{Zugner2019a}, and the results on Citeseer see~\autoref{sec:appendix_empiricalresults}.

\subsection{Certified robustness}\label{sec:experimentalresults}

In~\autoref{tab:results}, we summarize the certified robustness of the experiments on Cora ML and selected experiments on Citeseer. For a complete comparison, we also report the accuracy of the base classifier. In~\autoref{sec:appendix_fullresultstable}, we report results on all three datasets with error estimates. 
Our Soft Medoid GDC architecture comes with a relative increase on the accumulated certifications of more than 200\% w.r.t.\ adversarially added edges (most challenging case) for a wide range of baselines, alternative architectures, and defenses~\citep{Wu2019, Entezari2020, Zhu2019}. In the same scenario, on Citeseer we outperform the other baselines by a factor of 5.5. Moreover, our Soft Medoid GDC outperforms the ``hard'' Medoid as well as dimension-wise Median. As expected, increased robustness comes with the price of a slightly lower accuracy (compared to the best performing model which is substantially less robust).

\begin{wraptable}[22]{l}{.575\textwidth}
    \centering
    \vspace{-12pt}
    \captionof{table}{Accumulated certifications (first to third data column) and average certifiable radii (fourth and fifth data column) for the different architectures (top two highlighted). In the last column we list the clean accuracy of the base classifier (binary node attributes).}
    \label{tab:results}
    \resizebox{\linewidth}{!}{
    \begin{tabular}{llcccccc}
    \toprule
                                                  &         & \multicolumn{3}{l}{\textbf{Accum. certificates}} & \multicolumn{2}{l}{\textbf{Ave. cert. rad.}} &  \textbf{Acc.} \\
                                                  &   &              \textbf{A.\&d.} &      \textbf{Add} &     \textbf{Del.} &             \textbf{Add} & \multicolumn{2}{l}{\textbf{Del.}} \\
    \midrule
    \multirow{13}{*}{\rotatebox{90}{Cora ML~\citep{Bojchevski2018}}} & Vanilla GCN &                         1.84 &              0.21 &              4.42 &                     0.25 &              5.37 &              0.823 \\
                                                  & Vanilla GDC &                         1.98 &              0.20 &              4.33 &                     0.25 &              5.25 &  \underline{0.835} \\
                                                  & Vanilla APPNP &                         3.37 &              0.39 &              4.61 &                     0.47 &              5.53 &     \textbf{0.841} \\
                                                  & Vanilla GAT &                         1.26 &              0.07 &              4.03 &                     0.09 &              5.02 &              0.806 \\
                                                  & SVD GCN &                         0.84 &              0.08 &              2.39 &                     0.11 &              3.14 &              0.772 \\
                                                  & Jaccard GCN &                         0.86 &              0.01 &              4.39 &                     0.02 &              5.43 &              0.775 \\
                                                  & RGCN &                         1.46 &              0.12 &              3.99 &                     0.15 &              5.03 &              0.793 \\
                                                  & SM GCN ($T=50$) &                         1.86 &              0.21 &              4.44 &                     0.25 &              5.41 &              0.823 \\
                                                  & Dimmedian GDC &                         2.38 &              0.32 &              4.61 &                     0.41 &              5.71 &              0.801 \\
                                                  & Medoid GDC &                         4.05 &              0.51 &              4.62 &                     0.73 &  \underline{6.28} &              0.724 \\
                                                  & SM GDC ($T=1.0$) &                         4.31 &              0.52 &              4.71 &                     0.66 &              5.70 &              0.823 \\
                                                  & SM GDC ($T=0.5$) &             \underline{5.07} &  \underline{0.60} &  \underline{4.80} &         \underline{0.79} &              5.98 &              0.795 \\
                                                  & SM GDC ($T=0.2$) &                \textbf{5.60} &     \textbf{0.66} &     \textbf{4.91} &            \textbf{0.89} &     \textbf{6.31} &              0.770 \\
    \cline{1-8}
    \multirow{7}{*}{\rotatebox{90}{Citeseer~\citep{McCallum2000}}} & Vanilla GCN &                         1.24 &              0.11 &              3.88 &                     0.16 &              5.48 &              0.710 \\
                                                  & SVD GCN &                         0.52 &              0.00 &              2.12 &                     0.00 &              3.25 &              0.639 \\
                                                  & Jaccard GCN &                         1.42 &              0.04 &              3.96 &                     0.06 &              5.57 &  \underline{0.711} \\
                                                  & RGCN &                         1.12 &              0.09 &              3.89 &                     0.12 &              5.44 &     \textbf{0.719} \\
                                                  & SM GDC ($T=1.0$) &                         2.67 &              0.32 &              4.12 &                     0.45 &              5.77 &  \underline{0.711} \\
                                                  & SM GDC ($T=0.5$) &             \underline{3.62} &  \underline{0.48} &  \underline{4.22} &         \underline{0.69} &  \underline{5.94} &              0.709 \\
                                                  & SM GDC ($T=0.2$) &                \textbf{4.69} &     \textbf{0.60} &     \textbf{4.44} &            \textbf{0.89} &     \textbf{6.32} &              0.702 \\
    \bottomrule
    \end{tabular}
    }
\end{wraptable}

\textbf{Graph diffusion.} Node degrees in real-world graphs typically follow a power-law distribution. Consequently, we must be able to deal with a large fraction of low degree nodes. To obtain more robust GNNs, methods that are increasing the degree of the nodes are an important ingredient for the success of our model. The GDC architecture~\cite{Klicpera2019a} is one of the natural choices for smoothing the adjacency matrix because its low-pass filtering of the adjacency matrix leads to an increased number of non-zero weights.

To illustrate why the GDC architecture is well-suited for being equipped with the Soft Medoid, we plot the accumulated certifications over the degree in~\autoref{fig:accumcertsoverdegnogdc}. We see that with increasing degree the Soft Medoid GCN can demonstrate its strengths. We hypothesize, given just a few data points (i.e. neighbors), it is challenging for a robust estimator to differentiate between clean samples and outliers. Moreover, just a few adversarially added edges suffice to exceed the breakdown point. Note, however, that GDC alone does not improve the robustness by much (see~\autoref{fig:certratio} and \autoref{fig:accumcertsoverdegnogdc}). 

In conclusion of this discussion, the Soft Medoid and GDC synergize well and help to tackle the challenging problem of robustifying low degree nodes. In comparison to a GCN, with our approach, we can improve the robustness by up to eight times for low-degree nodes.

\textbf{Edge deletion.} For the case of edge deletion, a vanilla GCN performs already decently. This observation matches our experiments, where we found that with an identical budget it is more powerful to inject a few outliers than removing the same amount of ``good'' edges (in the sense of perturbing the message passing aggregation). 

\textbf{Attributes.} We observed that increased robustness against structure attacks comes with a decreased robustness on attribute attacks (GCN as baseline). Since we do not focus attribute robustness, we refer to~\autoref{sec:appendix_structurevsattrobust} for further insights and, at the same time, we present a parametrization of our approach which comes with improved attribute robustness.

\textbf{Defenses.} Complementary to~\autoref{tab:results}, in \autoref{fig:certratio}, we contrast the certification ratio for the Soft Medoid GDC to the state-of-the-art defenses~\citep{Entezari2020, Wu2019, Zhu2019} over different radii \(r_d\) and \(r_a\). Our model outperforms all of the tested state-of-the-art defenses by a large margin. All defenses~\citep{Wu2019, Entezari2020, Zhu2019} do not achieve high certification ratios. Thus, defenses designed for specific attacks cannot serve as general defenses against adversarial attacks. This highlights the need for certifiably robust models, as in general, we can make no a priori assumptions about adversarial attacks in the real world.

\begin{figure}
\centering
\resizebox{1.0\linewidth}{!}{
\begingroup%
\makeatletter%
\begin{pgfpicture}%
\pgfpathrectangle{\pgfpointorigin}{\pgfqpoint{6.456312in}{0.566613in}}%
\pgfusepath{use as bounding box, clip}%
\begin{pgfscope}%
\pgfsetbuttcap%
\pgfsetmiterjoin%
\definecolor{currentfill}{rgb}{1.000000,1.000000,1.000000}%
\pgfsetfillcolor{currentfill}%
\pgfsetlinewidth{0.000000pt}%
\definecolor{currentstroke}{rgb}{1.000000,1.000000,1.000000}%
\pgfsetstrokecolor{currentstroke}%
\pgfsetdash{}{0pt}%
\pgfpathmoveto{\pgfqpoint{0.000000in}{0.000000in}}%
\pgfpathlineto{\pgfqpoint{6.456312in}{0.000000in}}%
\pgfpathlineto{\pgfqpoint{6.456312in}{0.566612in}}%
\pgfpathlineto{\pgfqpoint{0.000000in}{0.566612in}}%
\pgfpathclose%
\pgfusepath{fill}%
\end{pgfscope}%
\begin{pgfscope}%
\pgfsetbuttcap%
\pgfsetmiterjoin%
\definecolor{currentfill}{rgb}{1.000000,1.000000,1.000000}%
\pgfsetfillcolor{currentfill}%
\pgfsetfillopacity{0.800000}%
\pgfsetlinewidth{1.003750pt}%
\definecolor{currentstroke}{rgb}{0.800000,0.800000,0.800000}%
\pgfsetstrokecolor{currentstroke}%
\pgfsetstrokeopacity{0.800000}%
\pgfsetdash{}{0pt}%
\pgfpathmoveto{\pgfqpoint{0.122222in}{0.100000in}}%
\pgfpathlineto{\pgfqpoint{6.334090in}{0.100000in}}%
\pgfpathquadraticcurveto{\pgfqpoint{6.356312in}{0.100000in}}{\pgfqpoint{6.356312in}{0.122222in}}%
\pgfpathlineto{\pgfqpoint{6.356312in}{0.444390in}}%
\pgfpathquadraticcurveto{\pgfqpoint{6.356312in}{0.466613in}}{\pgfqpoint{6.334090in}{0.466613in}}%
\pgfpathlineto{\pgfqpoint{0.122222in}{0.466613in}}%
\pgfpathquadraticcurveto{\pgfqpoint{0.100000in}{0.466613in}}{\pgfqpoint{0.100000in}{0.444390in}}%
\pgfpathlineto{\pgfqpoint{0.100000in}{0.122222in}}%
\pgfpathquadraticcurveto{\pgfqpoint{0.100000in}{0.100000in}}{\pgfqpoint{0.122222in}{0.100000in}}%
\pgfpathclose%
\pgfusepath{stroke,fill}%
\end{pgfscope}%
\begin{pgfscope}%
\pgfsetroundcap%
\pgfsetroundjoin%
\pgfsetlinewidth{1.003750pt}%
\definecolor{currentstroke}{rgb}{0.298039,0.447059,0.690196}%
\pgfsetstrokecolor{currentstroke}%
\pgfsetdash{}{0pt}%
\pgfpathmoveto{\pgfqpoint{0.144444in}{0.377744in}}%
\pgfpathlineto{\pgfqpoint{0.366667in}{0.377744in}}%
\pgfusepath{stroke}%
\end{pgfscope}%
\begin{pgfscope}%
\definecolor{textcolor}{rgb}{0.150000,0.150000,0.150000}%
\pgfsetstrokecolor{textcolor}%
\pgfsetfillcolor{textcolor}%
\pgftext[x=0.455556in,y=0.338855in,left,base]{\color{textcolor}\rmfamily\fontsize{8.000000}{9.600000}\selectfont Soft Medoid GDC (\(\displaystyle T=1.0\))}%
\end{pgfscope}%
\begin{pgfscope}%
\pgfsetroundcap%
\pgfsetroundjoin%
\pgfsetlinewidth{1.003750pt}%
\definecolor{currentstroke}{rgb}{0.866667,0.517647,0.321569}%
\pgfsetstrokecolor{currentstroke}%
\pgfsetdash{}{0pt}%
\pgfpathmoveto{\pgfqpoint{0.144444in}{0.211104in}}%
\pgfpathlineto{\pgfqpoint{0.366667in}{0.211104in}}%
\pgfusepath{stroke}%
\end{pgfscope}%
\begin{pgfscope}%
\definecolor{textcolor}{rgb}{0.150000,0.150000,0.150000}%
\pgfsetstrokecolor{textcolor}%
\pgfsetfillcolor{textcolor}%
\pgftext[x=0.455556in,y=0.172215in,left,base]{\color{textcolor}\rmfamily\fontsize{8.000000}{9.600000}\selectfont Soft Medoid GDC (\(\displaystyle T=0.2\))}%
\end{pgfscope}%
\begin{pgfscope}%
\pgfsetroundcap%
\pgfsetroundjoin%
\pgfsetlinewidth{1.003750pt}%
\definecolor{currentstroke}{rgb}{0.333333,0.658824,0.407843}%
\pgfsetstrokecolor{currentstroke}%
\pgfsetdash{}{0pt}%
\pgfpathmoveto{\pgfqpoint{2.138942in}{0.377744in}}%
\pgfpathlineto{\pgfqpoint{2.361165in}{0.377744in}}%
\pgfusepath{stroke}%
\end{pgfscope}%
\begin{pgfscope}%
\definecolor{textcolor}{rgb}{0.150000,0.150000,0.150000}%
\pgfsetstrokecolor{textcolor}%
\pgfsetfillcolor{textcolor}%
\pgftext[x=2.450054in,y=0.338855in,left,base]{\color{textcolor}\rmfamily\fontsize{8.000000}{9.600000}\selectfont Soft Medoid GCN (\(\displaystyle T=50\))}%
\end{pgfscope}%
\begin{pgfscope}%
\pgfsetbuttcap%
\pgfsetroundjoin%
\pgfsetlinewidth{1.003750pt}%
\definecolor{currentstroke}{rgb}{0.768627,0.305882,0.321569}%
\pgfsetstrokecolor{currentstroke}%
\pgfsetdash{{1.000000pt}{1.650000pt}}{0.000000pt}%
\pgfpathmoveto{\pgfqpoint{2.138942in}{0.216640in}}%
\pgfpathlineto{\pgfqpoint{2.361165in}{0.216640in}}%
\pgfusepath{stroke}%
\end{pgfscope}%
\begin{pgfscope}%
\definecolor{textcolor}{rgb}{0.150000,0.150000,0.150000}%
\pgfsetstrokecolor{textcolor}%
\pgfsetfillcolor{textcolor}%
\pgftext[x=2.450054in,y=0.177751in,left,base]{\color{textcolor}\rmfamily\fontsize{8.000000}{9.600000}\selectfont Vanilla GCN}%
\end{pgfscope}%
\begin{pgfscope}%
\pgfsetbuttcap%
\pgfsetroundjoin%
\pgfsetlinewidth{1.003750pt}%
\definecolor{currentstroke}{rgb}{0.505882,0.447059,0.701961}%
\pgfsetstrokecolor{currentstroke}%
\pgfsetdash{{1.000000pt}{1.650000pt}}{0.000000pt}%
\pgfpathmoveto{\pgfqpoint{4.098950in}{0.377744in}}%
\pgfpathlineto{\pgfqpoint{4.321172in}{0.377744in}}%
\pgfusepath{stroke}%
\end{pgfscope}%
\begin{pgfscope}%
\definecolor{textcolor}{rgb}{0.150000,0.150000,0.150000}%
\pgfsetstrokecolor{textcolor}%
\pgfsetfillcolor{textcolor}%
\pgftext[x=4.410061in,y=0.338855in,left,base]{\color{textcolor}\rmfamily\fontsize{8.000000}{9.600000}\selectfont Vanilla GDC}%
\end{pgfscope}%
\begin{pgfscope}%
\pgfsetbuttcap%
\pgfsetroundjoin%
\pgfsetlinewidth{1.003750pt}%
\definecolor{currentstroke}{rgb}{0.576471,0.470588,0.376471}%
\pgfsetstrokecolor{currentstroke}%
\pgfsetdash{{3.700000pt}{1.600000pt}}{0.000000pt}%
\pgfpathmoveto{\pgfqpoint{4.098950in}{0.222811in}}%
\pgfpathlineto{\pgfqpoint{4.321172in}{0.222811in}}%
\pgfusepath{stroke}%
\end{pgfscope}%
\begin{pgfscope}%
\definecolor{textcolor}{rgb}{0.150000,0.150000,0.150000}%
\pgfsetstrokecolor{textcolor}%
\pgfsetfillcolor{textcolor}%
\pgftext[x=4.410061in,y=0.183922in,left,base]{\color{textcolor}\rmfamily\fontsize{8.000000}{9.600000}\selectfont SVD GCN}%
\end{pgfscope}%
\begin{pgfscope}%
\pgfsetbuttcap%
\pgfsetroundjoin%
\pgfsetlinewidth{1.003750pt}%
\definecolor{currentstroke}{rgb}{0.854902,0.545098,0.764706}%
\pgfsetstrokecolor{currentstroke}%
\pgfsetdash{{3.700000pt}{1.600000pt}}{0.000000pt}%
\pgfpathmoveto{\pgfqpoint{5.300079in}{0.377744in}}%
\pgfpathlineto{\pgfqpoint{5.522301in}{0.377744in}}%
\pgfusepath{stroke}%
\end{pgfscope}%
\begin{pgfscope}%
\definecolor{textcolor}{rgb}{0.150000,0.150000,0.150000}%
\pgfsetstrokecolor{textcolor}%
\pgfsetfillcolor{textcolor}%
\pgftext[x=5.611190in,y=0.338855in,left,base]{\color{textcolor}\rmfamily\fontsize{8.000000}{9.600000}\selectfont Jaccard GCN}%
\end{pgfscope}%
\begin{pgfscope}%
\pgfsetbuttcap%
\pgfsetroundjoin%
\pgfsetlinewidth{1.003750pt}%
\definecolor{currentstroke}{rgb}{0.549020,0.549020,0.549020}%
\pgfsetstrokecolor{currentstroke}%
\pgfsetdash{{3.700000pt}{1.600000pt}}{0.000000pt}%
\pgfpathmoveto{\pgfqpoint{5.300079in}{0.222811in}}%
\pgfpathlineto{\pgfqpoint{5.522301in}{0.222811in}}%
\pgfusepath{stroke}%
\end{pgfscope}%
\begin{pgfscope}%
\definecolor{textcolor}{rgb}{0.150000,0.150000,0.150000}%
\pgfsetstrokecolor{textcolor}%
\pgfsetfillcolor{textcolor}%
\pgftext[x=5.611190in,y=0.183922in,left,base]{\color{textcolor}\rmfamily\fontsize{8.000000}{9.600000}\selectfont RGCN}%
\end{pgfscope}%
\end{pgfpicture}%
\makeatother%
\endgroup%
}
\\
\vspace{-10pt}
\begin{minipage}{.3\textwidth}
    \vspace{12pt}
  \centering
  \resizebox{\linewidth}{!}{\input{assets/cert_area_over_deg_gdc_no_legend_1.pgf}}
  \caption{Accumulated certifications (see~\autoref{eq:accumcert}) over the degree (equal freq. binning).\label{fig:accumcertsoverdegnogdc}}
\end{minipage}
\hspace{0.05cm}
\begin{minipage}{.65\textwidth}
  \centering
  \makebox[\columnwidth][c]{
    \(\begin{array}{cc} 
    \subfloat[]{\resizebox{0.49\linewidth}{!}{\input{assets/cert_ratio_gdc_del_no_legend_1.pgf}}} &
    \subfloat[]{\resizebox{0.4525\linewidth}{!}{\input{assets/cert_ratio_gdc_add_no_leglab_1.pgf}}}
    \end{array}\)
  }
  \caption{(a) and (b) show the certification ratio over different radii for deletion \(r_d\) and addition \(r_a\). We compare our the Soft Medoid GDC against a GCN and the other defenses~\citep{Entezari2020, Wu2019, Zhu2019}.\label{fig:certratio}}
\end{minipage}
\end{figure}

\section{Related work}\label{sec:relatedwork}

GNNs are an important class of deep neural networks, both from a scientific and application standpoint. Following the recent, trendsetting work in~\citep{Kipf2017, Hamilton2017}, a vast number of approaches were proposed~\citep{Gao2019a,Velickovic2018,AbuElHaija2019,Xu2019,Klicpera2019,Klicpera2019a}. A magnitude of adversarial attacks have been introduced~\citep{Dai2018, Zugner2018, Xu2019a, Zugner2019a, Bojchevski2019, Wu2019, Entezari2020, Miller2020}, pointing out their sensitivity regarding such attacks. Many of the proposed attacks directly propose an appropriate defense. We can classify the approaches into the categories of preprocessing~\citep{Entezari2020,Wu2019}, robust training~\citep{Xu2019a, Zugner2019a}, and modifications of the architecture~\citep{Zhu2019, Zhang2019a}. Perhaps the most similar approach, due to their statistical motivation, is RGCN~\citep{Zhu2019}.

An alternative direction to heuristic defenses is certification against small perturbations of the input~\cite{Hein2017, Wong2018}. Some of these ideas were transferred to GNNs, recently~\citep{Zuegner2020, Bojchevski2019a}. These certifications usually impose many restrictions regarding architectures or perturbations. In~\cite{Bojchevski2020}, randomized smoothing~\cite{Li2018, Lecuyer2019, Cohen2019} was extended to GNNs for an empirical certification of arbitrary architectures.

Note that our reasoning about robust location estimators is orthogonal to the work of \citet{Xu2019}. Their aim is to design aggregation functions that maximize the expressive power of a GNN. On the other hand, our goal is to design \emph{robust} aggregation functions. Since the 1960s, the robust statistics community has been systematically studying such estimators in the presence of outliers~\citep{Tukey1960, Huber1964}, and in recent years, research has also been drawn towards robust estimation in high dimensions~\citep{Diakonikolas2019b}.

\section{Conclusion}\label{sec:conclusion}

We propose a robust aggregation function, Soft Medoid, for the internal use within GNNs. We show that the Soft Medoid---a fully differentiable generalization of the Medoid---comes with the best possible breakdown point of 0.5 and an upper bound of the error/bias of the internal aggregations. 
We outperform all baseline and the other defenses~\citep{Wu2019, Entezari2020, Zhu2019} w.r.t.~robustness against structural perturbations by a relative margin of up to 450\% and for low-degree edges even 700\%. 

\section*{Broader Impact}\label{sec:broaderimpact}

This work is one step on the path towards the adversarial robustness of GNNs. Consequently, all potential applications of GNNs could benefit. These applications are computer vision, knowledge graphs, recommender systems, physics engines, and many more~\cite{Zhou2018, Wu2019a}.
Robust machine learning models certainly come with less opportunity of (fraudulent) manipulation. Robust models will enable the application of artificial intelligence (AI) for new use cases (e.g.\ safety-critical systems)---with all the related pros and cons. Perhaps, at some point, the discussion of risks and opportunities for AI~\citep{8456834, Balaram2018} and robust machine learning will converge.
Focusing on the negative aspects of contemporary applications, robust GNNs might cause, e.g., an increased automation bias~\cite{Mosier1998}, or fewer loopholes e.g.\ in the surveillance implemented in authoritarian systems~\cite{Report2018}. %

\begin{ack}
This research was supported by the German Research Foundation, Emmy Noether grant GU 1409/2-1, the German Federal Ministry of Education and Research (BMBF), grant no. 01IS18036B, and the Helmholtz Association under the joint research school ``Munich School for Data Science - MUDS.'' The authors of this work take full responsibilities for its content.
\end{ack}

\bibliography{references.bib}

\newpage
\appendix
\section{Proof of Soft Medoid breakdown point}\label{sec:appendix_proof_breakdown_sm}

Due to the extent of the proofs, we structure this section into three subsections. We start with a discussion of some preliminaries for the proofs. In~\autoref{sec:appendix_proof_lemma1}, we build upon~\citet{Lopuhaa1991}'s work to prove~\autoref{lemma:conditionapproxl1breakdown}. In~\autoref{sec:appendix_proof_softmedoid}, we prove~\autoref{theorem:softmedoidbreakdown}.

\subsection{Preliminaries}\label{sec:appendix_proof_prelim}

For this section, let \( \pertmset \) be decomposable such that \(\pertmset = \pertmset^{(\text{clean})} \cup \pertmset^{(\text{pert.})} \). Which is obtained from \(\featset = \{ \mathbf{\mathbf{x}}_1, \dots, \mathbf{\mathbf{x}}_n\} \), a collection of points in \(\mathbb{R}^d\), by replacing up to \(m\) points. The adversary can replace \(m\) arbitrary points. For a concise notation we simply write that we replace the first \(m\) values, but the points come with an arbitrary order beforehand. 

Note that our analysis is easily extendable to two further possible definitions of \(\pertmset\): (a) the adversary adds a perturbation to the original values \( \pertmset = \{ \mathbf{x}_1, \dots, \mathbf{x}_{m},\,\tilde{\mathbf{x}}_{m+1}, \dots, \tilde{\mathbf{x}}_{n} \} = \{ \mathbf{x}_{1} + \mathbf{p}_1, \dots, \mathbf{x}_{m} + \mathbf{p}_m, \mathbf{x}_{m+1}, \dots, \mathbf{x}_n \} \) (b) the adversary adds \(m\) perturbed samples to the dataset of \(n\) clean samples. In case (a), just some values during the derivation change, but the results are essentially the same. For case (b), the number of samples is no longer \(n\). Instead we have \(m+n\) samples and need to adjust the equations accordingly. In this case the estimator is not broken down if \(m < n\).

Recall that the Soft Medoid is orthogonal equivariant due to the Euclidean distance, i.e.\ \(t_{\text{SM}}(\mathbf{Q}\features + \mathbf{v}) = \mathbf{Q}\,t_{\text{SM}}(\features) + \mathbf{v}\), with the orthogonal matrix \(\mathbf{Q}\) and the translation vector \(\mathbf{v} \in \mathbb{R}^d\). \citet{Lopuhaa1991} proved for their Lemma 2.1 that a orthogonal equivariant location estimator \(t(\features)\) has the same breakdown point, regardless of how the clean data points were rotated and/or translated, i.e.\ \(\epsilon^*(t, \mathbf{A}\features + \mathbf{v}) = \epsilon^*(t, \features)\). For any nonsingular orthogonal matrix \(\mathbf{A}\) and translation vector \(\mathbf{v}\). This is a very powerful statement and allows us to simplify our setup significantly. We may transform the clean datapoints  such that \(t_{\text{SM}}(\features) = 0\) and rotate them arbitrarily.

\citet{Croux2002} derived a formula for the maxbias curve with Gaussian samples of an \(L_1\) estimator and pointed out that the worst case perturbation is a point mass. To sketch why this must also hold for the Soft Medoid, we analyze the ratio of the softmax output of a perturbed sample \(\hat{\softout}_{i}\) over the softmax output of a clean sample \(\hat{\softout}_{h}\):
\begin{equation}\label{eq:appendix_intutionofterms}
    \begin{aligned}
        \frac{\hat{\softout}_i}{\hat{\softout}_h} 
        &= \frac{\exp \Biggl\{ -\frac{1}{T} \Biggl[ \overbrace{\sum_{q \in \pertmset^{(\text{pert.})}} \|\tilde{\mathbf{x}}_q - \tilde{\mathbf{x}}_i\|}^{\alpha_1} + \overbrace{\sum_{o \in \pertmset^{(\text{clean)}}} \|\mathbf{x}_o - \tilde{\mathbf{x}}_i\|}^{\alpha_2} \Biggr] \Biggr\}}{\exp \Biggl \{-\frac{1}{T} \Biggl[ \underbrace{\sum_{q \in \pertmset^{(\text{pert.})}} \|\tilde{\mathbf{x}}_q - \mathbf{x}_h\|}_{\alpha_3} + \underbrace{\sum_{o \in \pertmset^{(\text{clean)}}} \|\mathbf{x}_o - \mathbf{x}_h\|}_{\alpha_4} \Biggr] \Biggr\}} \\
    \end{aligned}
\end{equation}
We see that we have four competing terms:
\begin{itemize}
    \item \(\alpha_1\) reflects some notion of variance of the perturbed samples (sum of distances from \(\tilde{\mathbf{x}}_i\) to all other \(\pertmset^{(\text{pert.})}\))
    \item \(\alpha_2\) is the sum of distances from the perturbed sample \(\tilde{\mathbf{x}}_i\) to all clean samples
    \item \(\alpha_3\) is the sum of distances from the clean sample \(\mathbf{x}_h\) to all perturbed samples
    \item \(\alpha_4\) reflects some notion of variance of the clean samples (sum of distances from \(\mathbf{x}_h\) to all other \(\pertmset^{(\text{clean})}\))
\end{itemize}
Note that if our goal was to maximize the influence of a perturbed samples, we would need to minimize \(\alpha_1\) and \(\alpha_2\). Analogously, we need would need to maximize \(\alpha_3\) and \(\alpha_4\). 

Hence, the bias benefits from a low variance of the perturbed samples \(\alpha_1\) and high variance of the clean samples \(\alpha_4\). The variance of the clean samples is something we typically cannot influence. \(\alpha_1\) clearly shows that the perturbed samples do not deviate in any way that does not help to maximize \(\nicefrac{\hat{\softout}_i}{\hat{\softout}_h}\). 
We even obtain the smallest possible value \(\alpha_1=0\), iff all perturbed samples coincide in one point. Moreover, \(\alpha_2\) and \(\alpha_3\) have a somewhat opposing objective. We can minimize \(\alpha_2\), by locating the perturbed sample \(\tilde{\mathbf{x}}_i\)  close to the clean samples. However, in this case, the clean estimate cannot be perturbed much.

Based on all these considerations, it is clear that it does not make sense that we have e.g.\ two groups of perturbed data points moving in opposite directions. For a \emph{worst case analysis} regarding perturbing the estimate towards infinity, we also must move \emph{all} perturbed data points towards infinity. Hence, we can simply model the worst-case perturbation as a point mass. %
Further, due to the orthogonal equivariance, we may assume w.l.o.g, for example, that the perturbation is located on the first axis since we can arbitrarily rotate the data beforehand. Additionally to the consideration of the location, for a perturbation approaching infinity, we are going to see that the solution is not depending on the absolute coordinates of the clean data points \(\featset\)

For solving several limits throughout our proofs, we make use of the following limit laws:
\begin{itemize}
    \item Law of addition: \(\lim_{x \to a} \left[f(x) + g(x)\right] = \lim_{x \to a}f(x) + \lim_{x \to a}g(x)\)
    \item Law of multiplication: \(\lim_{x \to a} \left[f(x) g(x)\right] = \left(\lim_{x \to a}f(x)\right) \left(\lim_{x \to a}g(x)\right)\)
    \item Law of division: \(\lim_{x \to a} \nicefrac{f(x)}{g(x)} =  \nicefrac{\lim_{x \to a}f(x)}{\lim_{x \to a}g(x)}\)\quad(if \(\lim_{x \to a}g(x) \ne 0\))
    \item Power law: \(\lim_{x \to a} \left[f(x)\right]^b = \left[\lim_{x \to a} f(x)\right]^b\)
    \item Composition law: \(\lim_{x \to a} f(g(x)) = f(\lim_{x \to a}g(x))\)\quad(if \(f(x)\) is continuous)
\end{itemize}
Further, note that a limit of a vector-valued function is evaluated element-wise. With \(\lim_{x \to \infty} f(x)\) we denote the limit towards positive infinity \(x \to +\infty\).

\subsection{Proof of~\autoref{lemma:conditionapproxl1breakdown}}\label{sec:appendix_proof_lemma1}

\begin{proof}\textit{Proof}~\autoref{lemma:conditionapproxl1breakdown}:
  Let \( \featset = \{\mathbf{x}_1, \dots, \mathbf{x}_n\} \) be the \(n\) clean data points. An adversary may replace an \( \epsilon \)-fraction (\( \epsilon = \nicefrac{m}{n} \)) and we denote the resulting set as \(\pertmset = \pertmset^{(\text{clean})} \cup \pertmset^{(\text{pert.})} \). \(\pertmset^{(\text{pert.})}  = \{\tilde{\mathbf{x}}_1, \dots, \tilde{\mathbf{x}}_m\} \) are the \(m\) replaced points and \(\pertmset^{(\text{clean})} \subset \featset \) the \(n-m\) remaining original data points. To ensure that \(\epsilon<0.5\), we may replace up to \(m = \lfloor\nicefrac{n-1}{2}\rfloor \) data points. Note that until~\autoref{eq:appendix_lemmacontraditction}, this proof is closely aligned to the ideas of~\citet{Lopuhaa1991}.
  
  We define \(M = \max_{\mathbf{x}_i \in \featset} \|\mathbf{x}_i\| \) and the ball \(B(\mathbf{0}, 2M)\) with its center in the origin and radius \(2M\). Moreover, we define the buffer \(b = \inf_{\mathbf{v} \in B(\mathbf{0}, 2M)} \le \|t(\pertm) - \mathbf{v}\| \). \(b\) denotes the minimum distance between \(t(\pertm)\) and \(B(\mathbf{0}, 2M)\) and from its definition \( \|t(\pertm)\| \le b + 2M\) follows. As a consequence for all \(m\) perturbed data points \( \tilde{\mathbf{x}}_q \in \pertmset^{(\text{pert.})} \) we have
  \begin{equation}\label{eq:appendix_lemmacond1}
    \| \tilde{\mathbf{x}}_q - t(\pertm) \| \ge \| \tilde{\mathbf{x}}_q \| -  \|t(\pertm) \|  \ge \| \tilde{\mathbf{x}}_q \| - (b + 2M)
  \end{equation} 

  Suppose that \(t(\pertm)\) and \(B(\mathbf{0}, 2M)\) are far from each other, i.e.\ \(b > 2 M m\). This definition seems to be arbitrary right now, but we are going to encounter exactly this term in~\autoref{eq:appendix_lemmacontraditction} and this assumption will lead to a contradiction if the condition of ~\autoref{lemma:conditionapproxl1breakdown} is satisfied. It is important to introduce this assumption right now, that we have for \( \mathbf{x}_o \in \pertmset^{(\text{clean})} \):
  \begin{equation}\label{eq:appendix_lemmacond2}
    \| \mathbf{x}_c - t(\pertm) \| \ge M + b \ge \| \mathbf{x}_o \| + b
  \end{equation}

  We now add~\autoref{eq:appendix_lemmacond1} and~\autoref{eq:appendix_lemmacond2} for all \(q\) and \(o\):
  \begin{equation}\label{eq:appendix_lemmacontraditction}
      \begin{aligned}
        \sum_{\tilde{\mathbf{x}}_j \in \pertmset} \| \tilde{\mathbf{x}}_j - t(\pertm) \| &\ge \left(\sum_{\tilde{\mathbf{x}}_q \in \pertmset^{(\text{pert.})}} \| \tilde{\mathbf{x}}_q \| - (b + 2M) \right) + \left( \sum_{\mathbf{x}_o \in \pertmset^{(\text{clean})}} \| \mathbf{x}_o \| + b \right) \\ &\ge \left(\sum_{\tilde{\mathbf{x}}_q \in \pertmset^{(\text{pert.})}} \| \tilde{\mathbf{x}}_q \| \right) - m b - 2 M m + \left( \sum_{\mathbf{x}_o \in \pertmset^{(\text{clean})}} \| \mathbf{x}_o \| \right) + (n-m) b  \\
        &\ge \left(\sum_{\tilde{\mathbf{x}}_i \in \pertmset} \| \tilde{\mathbf{x}} \|\right) + \underbrace{b - 2 M m}_{\begin{smallmatrix}> 0\\\text{(assumption)}\end{smallmatrix}} > \sum_{\tilde{\mathbf{x}}_i \in \pertmset} \| \tilde{\mathbf{x}} \|
      \end{aligned}
  \end{equation}
  Note that \(n-2m = n - 2\lfloor\nicefrac{n-1}{2}\rfloor \ge 1\). More precisely, if \(n\) is odd then \(n-2m = 1\), and if \(n\) is even then \(n-2m = 2\).
  
  Consequently, if we can show that
  \begin{equation}\label{eq:appendix_lemmacontraditctionrev}
      \sum_{\tilde{\mathbf{x}}_i \in \pertmset} \| \tilde{\mathbf{x}}_i - t(\pertm) \| \le \sum_{\tilde{\mathbf{x}}_i \in \pertmset} \| \tilde{\mathbf{x}} \|\,,
  \end{equation}
  or equivalently
  \begin{equation}\label{eq:appendix_lemmacontraditctionrev2}
      \frac{\sum_{\tilde{\mathbf{x}}_i \in \pertmset} \| \tilde{\mathbf{x}} - t(\pertm) \|}{\sum_{\tilde{\mathbf{x}}_i \in \pertmset} \| \tilde{\mathbf{x}} \|}  \le 1\,,
  \end{equation}
  we have a contradiction and, hence, \(b > 2 M m\) cannot be true. Similarly to~\cite{Lopuhaa1991}, this leads us to the worst case guarantee of \(\sup_{\pertm} \|t(\pertm) - t(\features) \| \le 2M(m+1) = 2\lfloor\nicefrac{n+1}{2}\rfloor M \le (n+1) \max_{\mathbf{x}_i \in \features} \|\mathbf{x}_i\|\). Please acknowledge that this rather loose guarantee holds for asymptotically \(\epsilon = 0.5\). This is the very worst case for which we can obtain a guarantee at all.
  
  In the last step towards~\autoref{lemma:conditionapproxl1breakdown} (\autoref{eq:appendix_lemmaperttoinfty} and \autoref{eq:appendix_lemmaperttoinfty2}), we analyze the case of a point mass perturbation on the first axis such that \( \pertmset = \{ \tilde{\mathbf{x}}_1, \dots, \tilde{\mathbf{x}}_m,\,\mathbf{x}_{m+1}, \dots,\mathbf{x}_{n} \} \) be the perturbed collection of points, with \(m = \lfloor \nicefrac{(n-1)}{2} \rfloor \) perturbed points \(\tilde{\mathbf{x}}_i = [\begin{matrix}p & 0 & \cdots & 0\end{matrix}]^\top, \forall i \in \{1, \dots, m\}\).
  
  The nature of the breakdown point definition (see~\autoref{eq:breakdown} in main part) requires us to show that the location estimate can approach infinity, once it has broken down. 
  Hence, the factors need to grow indefinitely, which they only can if the perturbation \(p\) approaches infinity. We are now going to analyze the resulting left side of
 ~\autoref{eq:appendix_lemmacontraditctionrev2}:
  
  \begin{equation}\label{eq:appendix_lemmaperttoinfty}
    \begin{aligned}
        & \lim_{p \to \infty} \frac{\sum_{\tilde{\mathbf{x}}_i \in \pertmset} \| \tilde{\mathbf{x}} - t(\pertm) \|}{\sum_{\tilde{\mathbf{x}}_i \in \pertmset} \| \tilde{\mathbf{x}} \|} \\
        &= \lim_{p \to \infty}
        \frac{\sum_{\tilde{\mathbf{x}}_q \in \pertmset^{(\text{pert.})}} \| \tilde{\mathbf{x}}_q - t(\pertm) \| + \sum_{\mathbf{x}_o \in \pertmset^{(\text{clean})}} \| \mathbf{x}_o - t(\pertm) \|}
        {\sum_{\tilde{\mathbf{x}}_q \in \pertmset^{(\text{pert.})}} \| \tilde{\mathbf{x}}_q \| + \sum_{\mathbf{x}_o \in \pertmset^{(\text{clean})}} \| \mathbf{x}_o \|} \\ 
        &= \lim_{p \to \infty} \frac{\frac{1}{p}}{\frac{1}{p}}
        \frac{\sum_{\tilde{\mathbf{x}}_q \in \pertmset^{(\text{pert.})}} \| \tilde{\mathbf{x}}_q - t(\pertm) \| + \sum_{\mathbf{x}_o \in \pertmset^{(\text{clean})}} \| \mathbf{x}_o - t(\pertm) \|}
        {\sum_{\tilde{\mathbf{x}}_q \in \pertmset^{(\text{pert.})}} \| \tilde{\mathbf{x}}_q \| + \sum_{\mathbf{x}_o \in \pertmset^{(\text{clean})}} \| \mathbf{x}_o \|} \\ 
        &= \lim_{p \to \infty}
        \frac{\sum_{\tilde{\mathbf{x}}_q \in \pertmset^{(\text{pert.})}} \| \frac{\tilde{\mathbf{x}}_q - t(\pertm)}{p} \| + \sum_{\mathbf{x}_o \in \pertmset^{(\text{clean})}} \| \frac{\mathbf{x}_o - t(\pertm)}{p} \|}
        {\sum_{\tilde{\mathbf{x}}_q \in \pertmset^{(\text{pert.})}} \| \frac{\tilde{\mathbf{x}}_q}{p} \| + \sum_{\mathbf{x}_o \in \pertmset^{(\text{clean})}} \| \frac{\mathbf{x}_o}{p} \|} \\ 
        &=
        \frac{\sum_{\tilde{\mathbf{x}}_q \in \pertmset^{(\text{pert.})}} \lim_{p \to \infty} \| \frac{\tilde{\mathbf{x}}_q - t(\pertm)}{p} \| + \sum_{\mathbf{x}_o \in \pertmset^{(\text{clean})}} \lim_{p \to \infty}\| \frac{\mathbf{x}_o - t(\pertm)}{p} \|}
        {\sum_{\tilde{\mathbf{x}}_q \in \pertmset^{(\text{pert.})}} \lim_{p \to \infty}\| \frac{\tilde{\mathbf{x}}_q}{p} \| + \sum_{\mathbf{x}_o \in \pertmset^{(\text{clean})}} \lim_{p \to \infty}\| \frac{\mathbf{x}_o}{p} \|} \\ 
        &=
        \frac{\sum\limits_{\tilde{\mathbf{x}}_q \in \pertmset^{(\text{pert.})}} \lim_{p \to \infty} \sqrt{\left(\frac{p - t(\pertm)_1}{p}\right)^2 + \sum\limits_{c=2}^d \left(\frac{t(\pertm)_c}{p}\right)^2} + \sum\limits_{\mathbf{x}_o \in \pertmset^{(\text{clean})}} \lim_{p \to \infty} \sqrt{\sum\limits_{c=1}^d \left(\frac{\mathbf{x}_{o,c} - t(\pertm)_c}{p}\right)^2}}
        {\sum\limits_{\tilde{\mathbf{x}}_q \in \pertmset^{(\text{pert.})}} 1} \\ 
        &= \frac{\sum\limits_{\tilde{\mathbf{x}}_q \in \pertmset^{(\text{pert.})}} \sqrt{\left(1 - \lim_{p \to \infty}\frac{ t(\pertm)_1}{p}\right)^2 + \sum\limits_{c=2}^d \left(\lim_{p \to \infty} \frac{t(\pertm)_c}{p}\right)^2} + \sum\limits_{\mathbf{x}_o \in \pertmset^{(\text{clean})}} \sqrt{\sum\limits_{c=1}^d \left(\lim_{p \to \infty} \frac{t(\pertm)_c}{p}\right)^2}}
        {|\pertmset^{(\text{pert.})}|} \\
    \end{aligned}
  \end{equation}
  
Since \(t(\pertm)\) is a vector-valued function, we denote its \(c\)-th component with \(t(\pertm)\) and, similarly, \(\mathbf{x}_{o,c}\) stands for the \(c\)-th component of vector \(\mathbf{x}_{o}\). Strictly speaking, the condition of~\autoref{lemma:conditionapproxl1breakdown} is not the only way to satisfy that~\autoref{eq:appendix_lemmaperttoinfty} is \(\le 1\). In the following, we focus on the most relevant case, though. If \( \lim_{p \to \infty} \nicefrac{t(\pertm)}{p} = \mathbf{0} \), \autoref{eq:appendix_lemmaperttoinfty} is \( \le 1\): 
  \begin{equation}\label{eq:appendix_lemmaperttoinfty2}
    \begin{aligned}
        \lim_{p \to \infty} \frac{\sum\limits_{\tilde{\mathbf{x}}_i \in \pertmset} \| \tilde{\mathbf{x}} - t(\pertm) \|}{\sum\limits_{\tilde{\mathbf{x}}_i \in \pertmset} \| \tilde{\mathbf{x}} \|}
        &= \frac{\sum\limits_{\tilde{\mathbf{x}}_q \in \pertmset^{(\text{pert.})}} \sqrt{\left(1 - 0\right)^2 + \sum\limits_{c=2}^d \left(0\right)^2} + \sum\limits_{\mathbf{x}_o \in \pertmset^{(\text{clean})}} \sqrt{\sum\limits_{c=1}^d \left(0\right)^2}}
        {|\pertmset^{(\text{pert.})}|}
        &= \frac{|\pertmset^{(\text{pert.})}|}{|\pertmset^{(\text{pert.})}|}
        &\le 1
    \end{aligned}
  \end{equation}
  Hence, the proof and~\autoref{lemma:conditionapproxl1breakdown} follows.
\end{proof}

\subsection{Breakdown point of the Soft Medoid}\label{sec:appendix_proof_softmedoid}

\begin{proof}\textit{Proof}~\autoref{theorem:softmedoidbreakdown}:
    Following up on~\autoref{lemma:conditionapproxl1breakdown}, we need to show that \( \lim_{p \to \infty}\nicefrac{t_{\text{SM}}(\pertm)}{p} = \lim_{p \to \infty} \nicefrac{1}{p}\,\hat{\softout}^\top\pertm = \mathbf{0}\). We can expand the limit of the vector-matrix multiplication \(\nicefrac{1}{p}\,\hat{\softout}^\top\pertm\) for its \(c\)-th component as (the result is going to be a vector):
    \begin{equation}\label{eq:appendix_limitvectormatrix}
        \left( \lim_{p \to \infty}\frac{\hat{\softout}^\top\pertm}{p} \right)_c = \lim_{p \to \infty} \sum_{i=1}^n \frac{\hat{\softout}_i \tilde{\mathbf{x}}_{i,c}}{p} = \sum_{i=1}^n \left(\lim_{p \to \infty} \hat{\softout}_i \right) \left(\lim_{p \to \infty} \frac{1}{p} \tilde{\mathbf{x}}_{i,c} \right) 
    \end{equation}
    
    From~\autoref{eq:appendix_limitvectormatrix} it is clear that we can take the element-wise limit of \(\lim_{p \to \infty} \hat{\softout}^\top\) and \(\lim_{p \to \infty} \nicefrac{1}{p}\,\pertm\), before the vector-matrix multiplication. The result is simply a sum of products of one element from the vector \(\softout\) and one of the matrix \(\nicefrac{1}{p}\,\pertm\). Hence, we are going to analyze \( \lim_{p \to \infty} \nicefrac{1}{p}\,\hat{\softout}^\top\pertm = \left(\lim_{p \to \infty} \hat{\softout}^\top \right) \left( \lim_{p \to \infty} \nicefrac{1}{p}\,\pertm\right) \).
    
    First, we analyze \(\lim_{p \to \infty} \nicefrac{\pertm}{p}\), i.e.\ the data matrix \( \pertm \) multiplied by the scalar factor of \(\nicefrac{1}{p}\):
    \begin{equation}\label{eq:appendix_matrixprooof}
        \lim_{p \to \infty} \frac{\pertm}{p} 
        = \lim_{p \to \infty} \frac{1}{p} \left[ \begin{matrix} p & 0 & \cdots & 0 \\ \vdots & \vdots & & \vdots \\ p & 0 & \cdots & 0 \\ \mathbf{x}_{m+1,1} & \mathbf{x}_{m+1,2} & \cdots & \mathbf{x}_{m+1,d} \\ \vdots & \vdots & & \vdots \\ \mathbf{x}_{n,1} & \mathbf{x}_{n,2} & \cdots & \mathbf{x}_{n,d}\end{matrix} \right] 
        = \left[ \begin{matrix} 1 & 0 & \cdots & 0 \\ \vdots & \vdots & & \vdots \\ 1 & 0 & \cdots & 0 \\ 0 & 0 & \cdots & 0 \\ \vdots & \vdots & & \vdots \\ 0 & 0 & \cdots & 0\end{matrix} \right]
    \end{equation}
    
    Second, since all weights in~\autoref{eq:appendix_matrixprooof} are zero, but for the first component of the perturbed samples, we solely need to show that the softmax of the perturbed samples is 0, i.e.\ \(\hat{\softout}_{q}=0\) for any \(q \in \pertmset^{(\text{pert.})}\). It is possible to show this directly. However, we go a different path to keep this proof brief. That is, analogously to the proof sketch of~\autoref{sec:breakdown}, the fraction of the softmax output of a perturbed sample \(\hat{\softout}_{q}\) over the softmax output of a clean sample \(\hat{\softout}_{c}=0\) (for \(c \in \pertmset^{(\text{clean})}\)) for \(p \to \infty\):
    
    \begin{equation}\label{eq:appendix_proofbdlaststep}
        \begin{aligned}
            \lim_{p \to \infty} \frac{\hat{\softout}_q}{\hat{\softout}_o} 
            &= \lim_{p \to \infty} \frac{\exp \left \{-\frac{1}{T} \sum_{\tilde{\mathbf{x}}_j \in \pertmset} \|\tilde{\mathbf{x}}_j - \tilde{\mathbf{x}}_q\| \right \}}{\exp \left \{-\frac{1}{T} \sum_{\tilde{\mathbf{x}}_j \in \pertmset} \|\tilde{\mathbf{x}}_j - \mathbf{x}_o\| \right \}} \\
            &= \lim_{p \to \infty} \underbrace{\frac{1}{\exp \left\{ -\frac{1}{T} \sum_{\mathbf{x}_i \in \pertmset^{(\text{clean})}} \|\mathbf{x}_i - \mathbf{x}_o\| \right\}}}_{\beta} \frac{\exp \left\{-\frac{1}{T}  \sum_{\tilde{\mathbf{x}}_i \in \pertmset^{(\text{clean})}} \|\mathbf{x}_j - \tilde{\mathbf{x}}_q\| \right\} }{ \exp \left\{ -\frac{1}{T} \sum_{\tilde{\mathbf{x}}_i \in \pertmset^{(\text{pert.})}} \|\tilde{\mathbf{x}}_j - \mathbf{x}_o\| \right\} } \\
            &= \beta \lim_{p \to \infty} \frac{\exp \left\{-\frac{1}{T}  \sum_{\tilde{\mathbf{x}}_i \in \pertmset^{(\text{clean})}} \|\mathbf{x}_j - \tilde{\mathbf{x}}_q\| \right\} }{ \exp \left\{ -\frac{1}{T} \sum_{\tilde{\mathbf{x}}_i \in \pertmset^{(\text{pert.})}} \|\tilde{\mathbf{x}}_j - \mathbf{x}_o\| \right\} } \\
            &= \beta \exp \left\{-\frac{1}{T} \left[ \sum_{\tilde{\mathbf{x}}_i \in \pertmset^{(\text{clean})}} \lim_{p \to \infty} \|\mathbf{x}_j - \tilde{\mathbf{x}}_q\| - \sum_{\tilde{\mathbf{x}}_i \in \pertmset^{(\text{pert.})}} \lim_{p \to \infty} \|\tilde{\mathbf{x}}_j - \mathbf{x}_o\| \right] \right\} \\
        \end{aligned}
    \end{equation}
    \begin{equation}\label{eq:appendix_proofbdlaststep2}
        \begin{aligned}
            \lim_{p \to \infty} \frac{\hat{\softout}_q}{\hat{\softout}_o} 
            &= \beta \exp \left\{-\frac{1}{T} \frac{\sum_{\tilde{\mathbf{x}}_i \in \pertmset^{(\text{clean})}} \lim_{p \to \infty} \|\frac{\mathbf{x}_j - \tilde{\mathbf{x}}_q}{p}\| - \sum_{\tilde{\mathbf{x}}_i \in \pertmset^{(\text{pert.})}} \lim_{p \to \infty} \|\frac{\tilde{\mathbf{x}}_j - \mathbf{x}_o}{p}\|}{\lim_{p \to \infty} \frac{1}{p}} \right\} \\
            &= \beta \exp \left\{-\frac{1}{T} \frac{\sum_{\tilde{\mathbf{x}}_i \in \pertmset^{(\text{clean})}} 1 - \sum_{\tilde{\mathbf{x}}_i \in \pertmset^{(\text{pert.})}} 1}{\frac{1}{\infty}} \right\} \\
            &= \beta \exp \left\{-\frac{1}{T} \frac{\overbrace{|\pertmset^{(\text{clean})}| - |\pertmset^{(\text{pert.})}|}^{> 0}}{\frac{1}{\infty}} \right\} \\
            &= \beta \exp \left\{- \infty \right\} \\
            &= 0 \\
        \end{aligned}
    \end{equation}
    
    Due to the range of the softmax \(\hat{\softout}_i \in [0,1], \forall i \in \{1, \dots, n\}\), a division by \(\infty\) in \(\nicefrac{\hat{\softout}_q}{\hat{\softout}_o}\) is not possible. It follows that \(\nicefrac{\hat{\softout}_q}{\hat{\softout}_o}=0\) iff the numerator \(\hat{\softout}_q = 0\). Consequently, \(\hat{\softout}_{q}=0, \forall q \in \pertmset^{(\text{pert.})}\) holds. This means that the weights for the perturbed samples approaches zero if \(p \to \infty\).
    
    In conclusion, the location estimate cannot approach infinity as long as the clean data points are finite and \( |\pertmset^{(\text{clean})}| > |\pertmset^{(\text{pert.})}|\), independent of the choice of \(T \in [0, \infty)\). 
\end{proof}

\subsection{Robustness of the Weighted Soft Medoid}\label{sec:appendix_proof_weightedsoftmedoid}

While the Weighted Soft Medoid's breakdown point holds for any positive weight vector, we present here quickly the results for the case of rational weights. In the context of a computer program, this is true anyways for numbers represented with finite precision. We only need to find the greatest common divisor of the weight vector \(\text{gcd}(\mathbf{a}) = \text{gcd}([\begin{matrix}\mathbf{a}_1 & \dots & \mathbf{a}_n \end{matrix}])\).

\begin{proof}\textit{Proof} weighted version of~\autoref{theorem:softmedoidbreakdown}: 
    We can transform \(\pertmset  = \{\tilde{\mathbf{x}}_1, \dots, \tilde{\mathbf{x}}_m, \mathbf{x}_{m+1}, \dots, \mathbf{x}_n \}\) with the according weight vector \(\mathbf{a} = [\begin{matrix}\mathbf{a}_1 & \dots & \mathbf{a}_n \end{matrix}]\) to its unweighted equivalent by this simple procedure: (1) we calculate \(\nicefrac{\mathbf{a}_j}{\text{gcd}(\mathbf{a})} = w_j,\, j \forall \{1, \dots, n\}\) and (2) we duplicate \(\tilde{\mathbf{x}}_j\) exactly \(w_j\) times and collect the results in the multiset \(\underline{\pertmset}\). Thereafter, we simply apply the unweighted Soft Medoid on the multiset \(\underline{\pertmset}\). It is apparent, that as long as
    \begin{equation}\label{eq:appendix_case_a}
        \sum_{q=1}^m \mathbf{a}_q < \sum_{o=m+1}^n \mathbf{a}_o
    \end{equation}
    holds, there are less perturbed examples than clean examples in \(\underline{\pertmset}\) and, hence, the estimator cannot be broken down.
\end{proof}

Note that the normalization etc.\ in~\autoref{eq:resulting-wsm}~(\autoref{sec:softmedoid}) does not affect the breakdown point of the estimator. In~\autoref{eq:appendix_proofbdlaststep2}, we show that the softmax weights for the perturbed samples approaches zero if \(p \to \infty\). These zero weights are still zero after the normalization; hence, the normalization does not influence the breakdown point.

\clearpage
\section{Detailed experimental results}\label{sec:appendix_detailed_results}

In this section, we present further and more detailed experimental results. We start with the dataset statistics in~\autoref{sec:appendix_datasets}. Thereafter, in~\autoref{sec:appendix_setupdetails}, we describe the setup regarding randomized smoothing in more detail. \autoref{sec:appendix_fullresultstable} summarizes the complete results of our experiments. In~\autoref{sec:appendix_structurevsattrobust}, we discuss the trade-off between robustness w.r.t.\ structural and attribute robustness. We conclude this section with further plots of the certification ratio over different radii.

\subsection{Dataset statistics}\label{sec:appendix_datasets}

For our experiments we use the largest connected component of the very common datasets summarized in~\autoref{tab:appendix_datasets}. In these citation graphs, the nodes of the graph represent publications and the edges citations. The node features are the one-hot encoding of the bag of words of the respective abstract. The classes of the semi-supervised prediction task represent different categories of the publications.

\begin{table}[ht]
  \centering
  \caption{Statistics of the largest connected component of the used datasets.}
  \label{tab:appendix_datasets}
  \resizebox{0.5\linewidth}{!}{
  \begin{tabular}{lrrr}
  \toprule
    {} & \textbf{\#Nodes $n$} & \textbf{\#Edges $e$} & \textbf{\#Features $d$} \\
    \midrule
    \textbf{Cora ML~\citep{Bojchevski2018}} &                2,810 &               15,962 &                   2,879 \\
    \textbf{Citeseer~\citep{McCallum2000}}  &                2,110 &                7,336 &                   3,703 \\
    \textbf{PubMed~\citep{Sen2008}}         &               19,717 &               88,648 &                     500 \\
    \bottomrule
  \end{tabular}
  }
\end{table}

\subsection{Randomized Smoothing}\label{sec:appendix_setupdetails}

In randomized smoothing, a deterministic or random base classifier \(f: \featset \to \mathcal{Y}\), that is a function from \(\mathbb{R}^d\) to the classes in \(\mathcal{Y}\), is extended to a smoothed classifier such that:
\begin{equation}\label{eq:appendix_randomizedsmoothing}
    g(\mathbf{x}) = \arg\max_{c \in \mathcal{Y}} \mathbb{P}(f(\phi(\mathbf{x})) = c)
\end{equation}
Where \(\phi(\mathbf{x})\) denotes some randomization around \(\mathbf{x}\). In the general case, there is not much hope to obtain the ensemble in closed form solution. Hence, in a Monte Carlo sampling setting, \(f(\phi(\mathbf{x}))\) is invoked multiple times. Then, the classification of the smooth ensemble is obtained via a majority vote among the different random inputs and the relative frequencies reflect class probabilities. \citet{Cohen2019} showed that for Gaussian noise one can obtain the certifiable \(L_2\)-ball radius depending on the difference between the most likely \(p_A\) and second most likely class \(p_B\). A certification of a radius \(r\) according to the \(L_2\)-ball means that with high probability (\(
1-\alpha_{\text{smoothing}}\)) the most likely class of the smooth classifier does not change if the perturbation of the input \(\delta\) is less or equal to the certified radius: \(\|\delta\| \le r\). To certify a large radius, we need (a) large difference \(p_A - p_B\), (b) a strong noise (e.g.\ high variance), and/or (c) many Monte Carlos samples of \(f(\phi(\mathbf{x}))\).

However, this definition of a certifiable radius does not reflect if the prediction was correct in the first place. Similarly to \citet{Cohen2019}, it makes sense to introduce a metric that combines correct prediction and certifiable robustness. This is why we report this conjunction throughout our experiments (see~\autoref{sec:expsetup}).

For the graph structure, we do have to deal with discrete values. Thus, applying Gaussian noise, as in~\citep{Cohen2019}, is not a good choice. \citet{Bojchevski2020} extended the framework of randomized smoothing to discrete variables in such a way that it is suitable for GNNs---considering the sparsity. For this purpose, they use a independent Bernoulli random variables that depend on the original data as randomization scheme \(\phi(\mathbf{x})\) and distinguish between a probability for deleting a binary feature or edge \(p_{-}\), as well as for adding a binary feature or edge \(p_{+}\). Lastly, we can obtain certificates at different radii for deletion \(r_d\) and addition \(r_a\) on the \(L_0\)-ball. Note that for real-world graphs we have much fewer edges than the nodes squared, which results in a sparse adjacency matrix \(\adj\) (most of the values are zero). If we used the same probability for adding and deleting edges, we would just delete a few edges but add comparatively many edges.

\citet{Cohen2019} argue that only a base classifier that is robust against the small perturbations \(\phi(\mathbf{x})\) can result in a certifiably robust smooth classifier. This is why we expect a base classifier that can be certified at a high radius to be more robust. For the smoothing we use the addition probability \(p_{+}=0.001\) and deletion probability \(p_{-}=0.4\), as suggested by~\cite{Bojchevski2020}. To obtain the certificates, we use a significance level of \(\alpha_{\text{smoothing}}=0.05\) and perform 10,000 forward passes.

\subsection{Empirical robustness}\label{sec:appendix_empiricalresults}

\begin{wraptable}[21]{r}{.5\textwidth}
\centering
\vspace{-12pt}
\caption{Targeted attack in the same setup as the evasion Nettack attack~\citep{Zugner2018}. We report the average margin and the failure rate of the attack (higher is better).}
\label{tab:appendix_nettack}
\resizebox{\linewidth}{!}{
    \begin{tabular}{llcc}
    \toprule
                                                  &                  & \multicolumn{2}{c}{\textbf{Nettack}} \\
                                                  &                  &                          \textbf{Margin} &                        \textbf{Fail. r.} \\
    \midrule
    \multirow{8}{*}{\rotatebox{90}{Cora ML~\citep{Bojchevski2018}}} & Vanilla GCN &                       -0.41 \(\pm\) 0.05 &                        0.18 \(\pm\) 0.06 \\
                                                  & Vanilla GDC &                       -0.48 \(\pm\) 0.12 &                        0.15 \(\pm\) 0.10 \\
                                                  & SVD GCN &            \underline{0.21 \(\pm\) 0.06} &  \textbf{0.64 \(\boldsymbol{\pm}\) 0.05} \\
                                                  & Jaccard GCN &                       -0.46 \(\pm\) 0.13 &                        0.26 \(\pm\) 0.07 \\
                                                  & RGCN &                        0.00 \(\pm\) 0.01 &                        0.35 \(\pm\) 0.01 \\
                                                  & SM GDC ($T=1.0$) &                        0.09 \(\pm\) 0.03 &                        0.54 \(\pm\) 0.02 \\
                                                  & SM GDC ($T=0.5$) &                        0.11 \(\pm\) 0.07 &                        0.53 \(\pm\) 0.08 \\
                                                  & SM GDC ($T=0.2$) &  \textbf{0.24 \(\boldsymbol{\pm}\) 0.06} &            \underline{0.62 \(\pm\) 0.04} \\
    \cline{1-4}
    \multirow{8}{*}{\rotatebox{90}{Citeseer~\citep{McCallum2000}}} & Vanilla GCN &                       -0.56 \(\pm\) 0.04 &                        0.03 \(\pm\) 0.00 \\
                                                  & Vanilla GDC &                       -0.51 \(\pm\) 0.02 &                        0.05 \(\pm\) 0.02 \\
                                                  & SVD GCN &                       -0.00 \(\pm\) 0.11 &            \underline{0.51 \(\pm\) 0.09} \\
                                                  & Jaccard GCN &                       -0.43 \(\pm\) 0.07 &                        0.17 \(\pm\) 0.07 \\
                                                  & RGCN &                       -0.05 \(\pm\) 0.03 &                        0.41 \(\pm\) 0.02 \\
                                                  & SM GDC ($T=1.0$) &                       -0.08 \(\pm\) 0.04 &                        0.36 \(\pm\) 0.04 \\
                                                  & SM GDC ($T=0.5$) &            \underline{0.10 \(\pm\) 0.07} &                        0.51 \(\pm\) 0.07 \\
                                                  & SM GDC ($T=0.2$) &  \textbf{0.31 \(\boldsymbol{\pm}\) 0.04} &  \textbf{0.65 \(\boldsymbol{\pm}\) 0.05} \\
    \bottomrule
    \end{tabular}
}
\end{wraptable}

For the empirical robustness (see~\autoref{sec:empirircalrobustness}), we use a surrogate GCN to perform the respective attack and adapt the adjacency matrix. We train the other models on the clean graph and only use the perturbed adjacency matrix for the prediction (evasion attack). Using the surrogate GCN comes with the main benefit that we do not evaluate how well a model might ``obfuscate'' the gradient towards the adjacency matrix. Moreover, every model has to face exactly the same changed edges.

\begin{figure}[h]
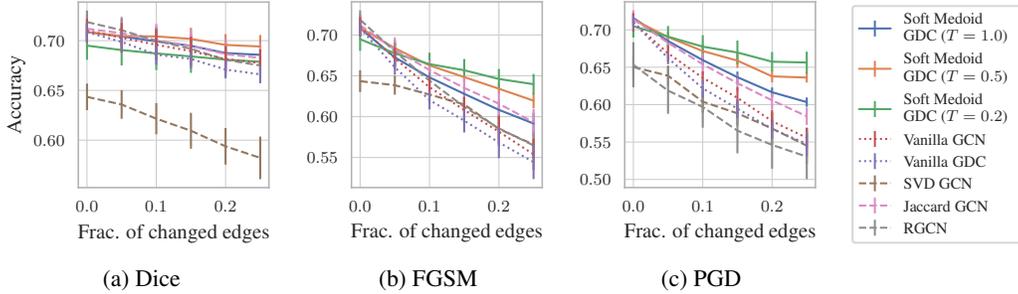

  \centering
  
  \makebox[\linewidth][c]{
    \(
    \arraycolsep=0pt
    \begin{array}{cccc} 
    \subfloat[Dice]{\resizebox{0.285\columnwidth}{!}{\input{assets/global_neurips_dice_citeseer_acc_no_legend.pgf}}} &
    \subfloat[FGSM]{\resizebox{0.26\columnwidth}{!}{\input{assets/global_neurips_fsgm_citeseer_acc_no_leglab.pgf}}} &
    \subfloat[PGD]{\resizebox{0.26\columnwidth}{!}{\input{assets/global_neurips_pgd_citeseer_acc_no_leglab.pgf}}} &
    \resizebox{0.19\columnwidth}{!}{\input{assets/global_neurips_fsgm_citeseer_acc_legend.pgf}}
    \end{array}
    \)
  }
  
  \caption{Accuracy for evasion (transfer) attacks on Citeseer.\label{fig:appendix_globalattack}}
\end{figure}

In~\autoref{fig:appendix_globalattack} we present the results on Citeseer in addition to~\autoref{fig:globalfsgm} of the main part. We see that our Soft Medoid outperforms the other approaches significantly for strong perturbations. In~\autoref{tab:appendix_nettack} we present the results on Nettack, where we outperform the other approaches as well. SVD GCN is the only exception and performs on par with our Soft Medoid GDC. Note that SVD GCN is specifically designed for Nettack. Last, \autoref{tab:appendix_global} complements \autoref{fig:globalfsgm} and \autoref{fig:appendix_globalattack} with selected numerical results including the standard error of the mean. Furthermore, the last two columns contain the results on the poisoning attack Metattack~\citep{Zugner2019a}. In conclusion, we see that our Soft Medoid GDC performs decently over a wide range of attacks. Having the results on certifiable robustness in mind (see~\autoref{sec:experimentalresults}), this comes at no surprise since certifiable robustness is an attack-agnostic measure of robustness.

\begin{table}[b]
\centering
\caption{Perturbed accuracy for the global attacks on Cora ML and Citeseer. Here \(\epsilon\) denotes the fraction of edges perturbed (relative to the clean graph).}
\label{tab:appendix_global}
\resizebox{\linewidth}{!}{
    \begin{tabular}{llcccccccc}
    \toprule
                                                  & \textbf{Attack} & \multicolumn{2}{c}{Dice} & \multicolumn{2}{c}{FGSM} & \multicolumn{2}{c}{PGD} & \multicolumn{2}{c}{Metattack} \\
                                                  & \textbf{Frac. pert. edges} $\boldsymbol{\epsilon}$ &                                       0.10 &                                       0.25 &                                       0.10 &                                       0.25 &                                       0.10 &                                       0.25 &                                       0.10 &                                       0.25 \\
    \midrule
    \multirow{8}{*}{\rotatebox{90}{Cora ML~\citep{Bojchevski2018}}} & Vanilla GCN &                        0.812 \(\pm\) 0.003 &                        0.785 \(\pm\) 0.004 &                        0.732 \(\pm\) 0.005 &                        0.655 \(\pm\) 0.003 &                        0.724 \(\pm\) 0.006 &                        0.619 \(\pm\) 0.006 &                        0.555 \(\pm\) 0.022 &                        0.383 \(\pm\) 0.012 \\
                                                  & Vanilla GDC &                        0.811 \(\pm\) 0.003 &                        0.789 \(\pm\) 0.001 &                        0.733 \(\pm\) 0.004 &                        0.657 \(\pm\) 0.003 &                        0.725 \(\pm\) 0.004 &                        0.624 \(\pm\) 0.007 &                        0.547 \(\pm\) 0.021 &                        0.342 \(\pm\) 0.012 \\
                                                  & SVD GCN &                        0.749 \(\pm\) 0.008 &                        0.710 \(\pm\) 0.008 &            \underline{0.750 \(\pm\) 0.007} &                        0.677 \(\pm\) 0.005 &                        0.736 \(\pm\) 0.009 &                        0.641 \(\pm\) 0.007 &  \textbf{0.699 \(\boldsymbol{\pm}\) 0.021} &  \textbf{0.496 \(\boldsymbol{\pm}\) 0.011} \\
                                                  & Jaccard GCN &                        0.801 \(\pm\) 0.002 &                        0.777 \(\pm\) 0.004 &                        0.733 \(\pm\) 0.002 &                        0.662 \(\pm\) 0.001 &                        0.722 \(\pm\) 0.003 &                        0.626 \(\pm\) 0.005 &                        0.584 \(\pm\) 0.023 &                        0.415 \(\pm\) 0.007 \\
                                                  & RGCN &                        0.782 \(\pm\) 0.000 &                        0.752 \(\pm\) 0.002 &                        0.721 \(\pm\) 0.002 &                        0.647 \(\pm\) 0.004 &                        0.712 \(\pm\) 0.006 &                        0.613 \(\pm\) 0.006 &                        0.591 \(\pm\) 0.028 &                        0.359 \(\pm\) 0.012 \\
                                                  & SM GDC ($T=1.0$) &  \textbf{0.818 \(\boldsymbol{\pm}\) 0.005} &  \textbf{0.801 \(\boldsymbol{\pm}\) 0.003} &                        0.743 \(\pm\) 0.001 &                        0.679 \(\pm\) 0.003 &                        0.739 \(\pm\) 0.002 &                        0.656 \(\pm\) 0.002 &                        0.608 \(\pm\) 0.028 &                        0.433 \(\pm\) 0.013 \\
                                                  & SM GDC ($T=0.5$) &            \underline{0.813 \(\pm\) 0.003} &            \underline{0.796 \(\pm\) 0.004} &  \textbf{0.751 \(\boldsymbol{\pm}\) 0.001} &            \underline{0.693 \(\pm\) 0.002} &            \underline{0.750 \(\pm\) 0.003} &            \underline{0.680 \(\pm\) 0.003} &                        0.626 \(\pm\) 0.024 &                        0.459 \(\pm\) 0.024 \\
                                                  & SM GDC ($T=0.2$) &                        0.787 \(\pm\) 0.002 &                        0.777 \(\pm\) 0.004 &                        0.749 \(\pm\) 0.000 &  \textbf{0.702 \(\boldsymbol{\pm}\) 0.001} &  \textbf{0.755 \(\boldsymbol{\pm}\) 0.002} &  \textbf{0.719 \(\boldsymbol{\pm}\) 0.002} &            \underline{0.641 \(\pm\) 0.026} &            \underline{0.474 \(\pm\) 0.029} \\
    \cline{1-10}
    \multirow{8}{*}{\rotatebox{90}{Citeseer~\citep{McCallum2000}}} & Vanilla GCN &                        0.696 \(\pm\) 0.016 &                        0.678 \(\pm\) 0.014 &                        0.642 \(\pm\) 0.014 &                        0.570 \(\pm\) 0.022 &                        0.636 \(\pm\) 0.009 &                        0.556 \(\pm\) 0.013 &                        0.587 \(\pm\) 0.021 &                        0.439 \(\pm\) 0.035 \\
                                                  & Vanilla GDC &                        0.686 \(\pm\) 0.010 &                        0.666 \(\pm\) 0.009 &                        0.635 \(\pm\) 0.011 &                        0.563 \(\pm\) 0.022 &                        0.622 \(\pm\) 0.012 &                        0.548 \(\pm\) 0.017 &                        0.598 \(\pm\) 0.017 &                        0.450 \(\pm\) 0.024 \\
                                                  & SVD GCN &                        0.622 \(\pm\) 0.016 &                        0.582 \(\pm\) 0.021 &                        0.625 \(\pm\) 0.016 &                        0.566 \(\pm\) 0.022 &                        0.604 \(\pm\) 0.015 &                        0.545 \(\pm\) 0.024 &  \textbf{0.631 \(\boldsymbol{\pm}\) 0.019} &  \textbf{0.531 \(\boldsymbol{\pm}\) 0.044} \\
                                                  & Jaccard GCN &            \underline{0.700 \(\pm\) 0.017} &                        0.683 \(\pm\) 0.015 &                        0.659 \(\pm\) 0.013 &                        0.601 \(\pm\) 0.016 &                        0.654 \(\pm\) 0.014 &                        0.584 \(\pm\) 0.012 &            \underline{0.620 \(\pm\) 0.019} &                        0.503 \(\pm\) 0.035 \\
                                                  & RGCN &                        0.700 \(\pm\) 0.013 &                        0.675 \(\pm\) 0.009 &                        0.593 \(\pm\) 0.030 &                        0.536 \(\pm\) 0.029 &                        0.597 \(\pm\) 0.028 &                        0.530 \(\pm\) 0.030 &                        0.615 \(\pm\) 0.014 &                        0.500 \(\pm\) 0.041 \\
                                                  & SM GDC ($T=1.0$) &                        0.699 \(\pm\) 0.012 &            \underline{0.686 \(\pm\) 0.012} &                        0.664 \(\pm\) 0.009 &                        0.606 \(\pm\) 0.012 &                        0.660 \(\pm\) 0.005 &                        0.603 \(\pm\) 0.006 &                        0.617 \(\pm\) 0.004 &                        0.502 \(\pm\) 0.033 \\
                                                  & SM GDC ($T=0.5$) &  \textbf{0.704 \(\boldsymbol{\pm}\) 0.011} &  \textbf{0.694 \(\boldsymbol{\pm}\) 0.012} &            \underline{0.674 \(\pm\) 0.009} &            \underline{0.631 \(\pm\) 0.012} &            \underline{0.672 \(\pm\) 0.009} &            \underline{0.636 \(\pm\) 0.007} &                        0.612 \(\pm\) 0.006 &                        0.506 \(\pm\) 0.028 \\
                                                  & SM GDC ($T=0.2$) &                        0.687 \(\pm\) 0.017 &                        0.679 \(\pm\) 0.017 &  \textbf{0.682 \(\boldsymbol{\pm}\) 0.013} &  \textbf{0.649 \(\boldsymbol{\pm}\) 0.012} &  \textbf{0.678 \(\boldsymbol{\pm}\) 0.015} &  \textbf{0.656 \(\boldsymbol{\pm}\) 0.015} &                        0.613 \(\pm\) 0.004 &            \underline{0.512 \(\pm\) 0.013} \\
    \bottomrule
    \end{tabular}
}
\end{table}

\subsection{Certified robustness}\label{sec:appendix_fullresultstable}

\autoref{tab:appendix_results} presents the complete results on CoraML, Citeseer as well as PubMed of our experiments with three-sigma error of the mean. On all three datasets Cora ML, Citeseer, and PubMed, our Soft Medoid GDC improves the robustness significantly. One of our Soft Medoid GDC models is in every structure robustness benchmark the most robust model. Moreover, we see that the the accuracy of the base classifier and the smooth classifier barely differ. We refer to~\autoref{sec:appendix_structurevsattrobust} for our Soft Medoid\(^\dagger\) GDC (\(T=10\)) model, which improves the attribute robustness.

\begin{table}
\centering
\caption{Summary of accumulated certifications and accuracy for the different architectures on Cora ML and Citeseer. We also report the accuracy of the base and smooth classifier (binary attr.).}
\label{tab:appendix_results}
\resizebox{\linewidth}{!}{
\begin{tabular}{llcccccc}
\toprule
                                       &         &                           \textbf{Attr.} & \multicolumn{3}{c}{\textbf{Edges}} &       \textbf{Accuracy (base)} &     \textbf{Accuracy (smooth)} \\
                                       &   &                          \textbf{A.\&d.} &                          \textbf{A.\&d.} &                             \textbf{Add} & \textbf{Del.} & & \\
\midrule
\multirow{14}{*}{\rotatebox{90}{Cora ML~\citep{Bojchevski2018}}} & Vanilla GCN &                        5.73 \(\pm\) 0.23 &                        1.84 \(\pm\) 0.01 &                        0.21 \(\pm\) 0.00 &                        4.42 \(\pm\) 0.01 &              0.823 $\pm$ 0.006 &              0.816 $\pm$ 0.006 \\
                                       & Vanilla GDC &                        5.80 \(\pm\) 0.06 &                        1.98 \(\pm\) 0.04 &                        0.20 \(\pm\) 0.00 &                        4.33 \(\pm\) 0.02 &  \underline{0.825 $\pm$ 0.007} &              0.824 $\pm$ 0.007 \\
                                       & Vanilla APPNP &                        5.57 \(\pm\) 0.04 &                        3.37 \(\pm\) 0.02 &                        0.39 \(\pm\) 0.01 &                        4.61 \(\pm\) 0.00 &     \textbf{0.836 $\pm$ 0.008} &     \textbf{0.837 $\pm$ 0.008} \\
                                       & Vanilla GAT &            \underline{5.83 \(\pm\) 0.09} &                        1.26 \(\pm\) 0.09 &                        0.07 \(\pm\) 0.01 &                        4.03 \(\pm\) 0.07 &              0.804 $\pm$ 0.002 &              0.807 $\pm$ 0.004 \\
                                       & SVD GCN &                        5.51 \(\pm\) 0.14 &                        0.84 \(\pm\) 0.09 &                        0.08 \(\pm\) 0.02 &                        2.39 \(\pm\) 0.04 &              0.772 $\pm$ 0.008 &              0.772 $\pm$ 0.007 \\
                                       & Jaccard GCN &                        5.59 \(\pm\) 0.15 &                        0.86 \(\pm\) 0.10 &                        0.01 \(\pm\) 0.01 &                        4.39 \(\pm\) 0.00 &              0.777 $\pm$ 0.003 &              0.778 $\pm$ 0.003 \\
                                       & RGCN &                        4.64 \(\pm\) 0.07 &                        1.46 \(\pm\) 0.03 &                        0.12 \(\pm\) 0.01 &                        3.99 \(\pm\) 0.08 &              0.796 $\pm$ 0.007 &              0.802 $\pm$ 0.005 \\
                                       & SM GCN ($T=50$) &                        5.68 \(\pm\) 0.05 &                        1.86 \(\pm\) 0.03 &                        0.21 \(\pm\) 0.00 &                        4.44 \(\pm\) 0.02 &              0.823 $\pm$ 0.003 &  \underline{0.825 $\pm$ 0.003} \\
                                       & Dimmedian GDC &                        4.66 \(\pm\) 0.05 &                        2.38 \(\pm\) 0.05 &                        0.32 \(\pm\) 0.01 &                        4.61 \(\pm\) 0.03 &              0.804 $\pm$ 0.002 &              0.805 $\pm$ 0.001 \\
                                       & Medoid GDC &                        1.98 \(\pm\) 0.07 &                        4.05 \(\pm\) 0.15 &                        0.51 \(\pm\) 0.02 &                        4.62 \(\pm\) 0.06 &              0.742 $\pm$ 0.008 &              0.756 $\pm$ 0.011 \\
                                       & SM GDC ($T=1.0$) &                        5.07 \(\pm\) 0.46 &                        4.31 \(\pm\) 0.68 &                        0.52 \(\pm\) 0.09 &                        4.71 \(\pm\) 0.08 &              0.819 $\pm$ 0.008 &              0.822 $\pm$ 0.007 \\
                                       & SM GDC ($T=0.5$) &                        4.15 \(\pm\) 0.67 &            \underline{5.07 \(\pm\) 0.74} &            \underline{0.60 \(\pm\) 0.08} &            \underline{4.80 \(\pm\) 0.07} &              0.796 $\pm$ 0.010 &              0.803 $\pm$ 0.008 \\
                                       & SM GDC ($T=0.2$) &                        2.90 \(\pm\) 0.95 &  \textbf{5.60 \(\boldsymbol{\pm}\) 0.31} &  \textbf{0.66 \(\boldsymbol{\pm}\) 0.04} &  \textbf{4.91 \(\boldsymbol{\pm}\) 0.04} &              0.768 $\pm$ 0.033 &              0.775 $\pm$ 0.034 \\
                                       & SM GDC$^\dagger$ ($T=10$) &  \textbf{7.15 \(\boldsymbol{\pm}\) 0.01} &                        1.12 \(\pm\) 0.06 &                        0.10 \(\pm\) 0.00 &                        1.63 \(\pm\) 0.01 &              0.811 $\pm$ 0.003 &              0.814 $\pm$ 0.002 \\
\cline{1-8}
\multirow{14}{*}{\rotatebox{90}{Citeseer~\citep{McCallum2000}}} & Vanilla GCN &                        4.43 \(\pm\) 0.21 &                        1.24 \(\pm\) 0.10 &                        0.11 \(\pm\) 0.01 &                        3.88 \(\pm\) 0.17 &              0.712 $\pm$ 0.008 &              0.712 $\pm$ 0.009 \\
                                       & Vanilla GDC &            \underline{5.21 \(\pm\) 0.22} &                        1.13 \(\pm\) 0.10 &                        0.09 \(\pm\) 0.01 &                        3.85 \(\pm\) 0.13 &              0.703 $\pm$ 0.007 &              0.701 $\pm$ 0.007 \\
                                       & Vanilla APPNP &                        5.08 \(\pm\) 0.04 &                        2.21 \(\pm\) 0.06 &                        0.23 \(\pm\) 0.01 &                        4.16 \(\pm\) 0.04 &  \underline{0.724 $\pm$ 0.005} &  \underline{0.723 $\pm$ 0.004} \\
                                       & Vanilla GAT &                        3.60 \(\pm\) 0.34 &                        0.66 \(\pm\) 0.13 &                        0.02 \(\pm\) 0.01 &                        3.24 \(\pm\) 0.48 &              0.652 $\pm$ 0.034 &              0.634 $\pm$ 0.044 \\
                                       & SVD GCN &                        3.46 \(\pm\) 0.13 &                        0.52 \(\pm\) 0.11 &                        0.00 \(\pm\) 0.00 &                        2.12 \(\pm\) 0.07 &              0.638 $\pm$ 0.015 &              0.634 $\pm$ 0.016 \\
                                       & Jaccard GCN &                        3.09 \(\pm\) 0.19 &                        1.42 \(\pm\) 0.10 &                        0.04 \(\pm\) 0.04 &                        3.96 \(\pm\) 0.14 &              0.711 $\pm$ 0.013 &              0.712 $\pm$ 0.012 \\
                                       & RGCN &                        4.27 \(\pm\) 0.18 &                        1.12 \(\pm\) 0.05 &                        0.09 \(\pm\) 0.01 &                        3.89 \(\pm\) 0.11 &              0.719 $\pm$ 0.012 &              0.718 $\pm$ 0.009 \\
                                       & SM GCN ($T=50$) &                        4.40 \(\pm\) 0.26 &                        1.25 \(\pm\) 0.10 &                        0.11 \(\pm\) 0.01 &                        3.90 \(\pm\) 0.17 &              0.711 $\pm$ 0.012 &              0.710 $\pm$ 0.013 \\
                                       & Dimmedian GDC &                        4.28 \(\pm\) 0.14 &                        1.42 \(\pm\) 0.05 &                        0.15 \(\pm\) 0.01 &                        3.92 \(\pm\) 0.08 &     \textbf{0.725 $\pm$ 0.012} &     \textbf{0.725 $\pm$ 0.011} \\
                                       & Medoid GDC &                        1.69 \(\pm\) 0.13 &                        2.41 \(\pm\) 0.04 &                        0.24 \(\pm\) 0.01 &                        3.97 \(\pm\) 0.06 &              0.673 $\pm$ 0.012 &              0.689 $\pm$ 0.007 \\
                                       & SM GDC ($T=1.0$) &                        4.93 \(\pm\) 0.24 &                        2.67 \(\pm\) 0.07 &                        0.32 \(\pm\) 0.02 &                        4.12 \(\pm\) 0.09 &              0.711 $\pm$ 0.010 &              0.712 $\pm$ 0.010 \\
                                       & SM GDC ($T=0.5$) &                        4.55 \(\pm\) 0.16 &            \underline{3.62 \(\pm\) 0.19} &            \underline{0.48 \(\pm\) 0.03} &            \underline{4.22 \(\pm\) 0.12} &              0.709 $\pm$ 0.010 &              0.716 $\pm$ 0.010 \\
                                       & SM GDC ($T=0.2$) &                        3.52 \(\pm\) 0.17 &  \textbf{4.69 \(\boldsymbol{\pm}\) 0.20} &  \textbf{0.60 \(\boldsymbol{\pm}\) 0.02} &  \textbf{4.44 \(\boldsymbol{\pm}\) 0.13} &              0.705 $\pm$ 0.017 &              0.714 $\pm$ 0.014 \\
                                       & SM GDC$^\dagger$ ($T=10$) &  \textbf{5.62 \(\boldsymbol{\pm}\) 0.15} &                        0.17 \(\pm\) 0.02 &                        0.02 \(\pm\) 0.00 &                        0.82 \(\pm\) 0.12 &              0.663 $\pm$ 0.014 &              0.654 $\pm$ 0.014 \\
\cline{1-8}
\multirow{5}{*}{\rotatebox{90}{PubMed~\citep{Sen2008}}} & Vanilla GCN &  \textbf{4.40 \(\boldsymbol{\pm}\) 0.20} &                        3.23 \(\pm\) 0.17 &                        0.22 \(\pm\) 0.02 &                        4.19 \(\pm\) 0.06 &              0.760 $\pm$ 0.026 &              0.744 $\pm$ 0.031 \\
                                       & Vanilla GDC &            \underline{4.32 \(\pm\) 0.11} &                        3.10 \(\pm\) 0.04 &                        0.24 \(\pm\) 0.01 &                        4.05 \(\pm\) 0.15 &     \textbf{0.764 $\pm$ 0.034} &              0.749 $\pm$ 0.039 \\
                                       & SM GDC ($T=1.0$) &                        3.30 \(\pm\) 0.41 &                        5.13 \(\pm\) 0.66 &                        0.47 \(\pm\) 0.04 &                        4.24 \(\pm\) 0.16 &  \underline{0.761 $\pm$ 0.023} &     \textbf{0.756 $\pm$ 0.027} \\
                                       & SM GDC ($T=0.5$) &                        3.10 \(\pm\) 0.60 &            \underline{5.43 \(\pm\) 0.26} &            \underline{0.56 \(\pm\) 0.03} &            \underline{4.35 \(\pm\) 0.16} &              0.751 $\pm$ 0.015 &  \underline{0.752 $\pm$ 0.019} \\
                                       & SM GDC ($T=0.2$) &                        2.44 \(\pm\) 0.40 &  \textbf{6.07 \(\boldsymbol{\pm}\) 0.19} &  \textbf{0.66 \(\boldsymbol{\pm}\) 0.02} &  \textbf{4.46 \(\boldsymbol{\pm}\) 0.15} &              0.729 $\pm$ 0.014 &              0.732 $\pm$ 0.016 \\
\bottomrule
\end{tabular}
}
\end{table}

On Pubmed, due to the runtime, we do not report the results of the other defenses, select some important baselines. In their original papers, RGCN~\cite{Zhu2019} is the only other defense~\cite{Wu2019, Entezari2020, Zhu2019} that reports results on a bigger data set, such as PubMed.

\subsection{Structural vs. attribute robustness}\label{sec:appendix_structurevsattrobust}

It is noticeable in~\autoref{tab:appendix_results} that increased robustness against structure attacks comes with a decreased robustness on attribute attacks (GCN as the baseline). This finding seems to be very consistent regardless of the chosen approach. For example, GAT outperforms APPNP on attribute attacks but lags behind APPNP on structure attacks. 

For an increased robustness against attribute attacks, we came up with an alternative normalization
\begin{equation}\label{eq:resulting-wsm-alternative}
  t^\dagger_{\text{WSM}}(\features, \mathbf{a}, T) 
  = \left(\sum_{i=1}^n \mathbf \adj_{v,i} \right) \softout^T \features
\end{equation}
of our Soft Medoid estimator and different choice of temperature \(T=10\). Note that the Soft Medoid does not have the GCN as a special case~\autoref{eq:resulting-wsm}~(\autoref{sec:softmedoid}). This configuration comes with the highest attribute robustness of all tested architectures (about 15\% to 30\% higher accumulated certifications w.r.t. attribute attacks than a GCN). For a comparison of the precise results see SM\(^\dagger\) GDC (\(T=10\)) in~\autoref{tab:appendix_results}.

\subsection{Detailed comparison of certification ratios}\label{sec:appendix_certratiocomparison}

\begin{figure}[H]
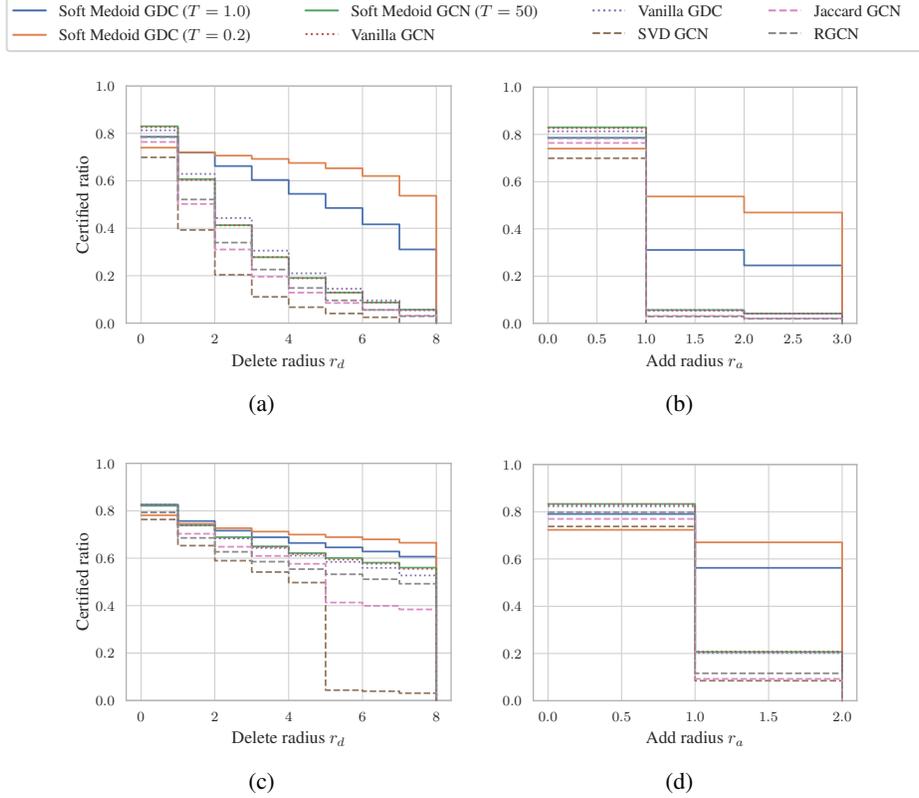

\centering
\resizebox{0.9\linewidth}{!}{
\begingroup%
\makeatletter%
\begin{pgfpicture}%
\pgfpathrectangle{\pgfpointorigin}{\pgfqpoint{6.456312in}{0.566613in}}%
\pgfusepath{use as bounding box, clip}%
\begin{pgfscope}%
\pgfsetbuttcap%
\pgfsetmiterjoin%
\definecolor{currentfill}{rgb}{1.000000,1.000000,1.000000}%
\pgfsetfillcolor{currentfill}%
\pgfsetlinewidth{0.000000pt}%
\definecolor{currentstroke}{rgb}{1.000000,1.000000,1.000000}%
\pgfsetstrokecolor{currentstroke}%
\pgfsetdash{}{0pt}%
\pgfpathmoveto{\pgfqpoint{0.000000in}{0.000000in}}%
\pgfpathlineto{\pgfqpoint{6.456312in}{0.000000in}}%
\pgfpathlineto{\pgfqpoint{6.456312in}{0.566612in}}%
\pgfpathlineto{\pgfqpoint{0.000000in}{0.566612in}}%
\pgfpathclose%
\pgfusepath{fill}%
\end{pgfscope}%
\begin{pgfscope}%
\pgfsetbuttcap%
\pgfsetmiterjoin%
\definecolor{currentfill}{rgb}{1.000000,1.000000,1.000000}%
\pgfsetfillcolor{currentfill}%
\pgfsetfillopacity{0.800000}%
\pgfsetlinewidth{1.003750pt}%
\definecolor{currentstroke}{rgb}{0.800000,0.800000,0.800000}%
\pgfsetstrokecolor{currentstroke}%
\pgfsetstrokeopacity{0.800000}%
\pgfsetdash{}{0pt}%
\pgfpathmoveto{\pgfqpoint{0.122222in}{0.100000in}}%
\pgfpathlineto{\pgfqpoint{6.334090in}{0.100000in}}%
\pgfpathquadraticcurveto{\pgfqpoint{6.356312in}{0.100000in}}{\pgfqpoint{6.356312in}{0.122222in}}%
\pgfpathlineto{\pgfqpoint{6.356312in}{0.444390in}}%
\pgfpathquadraticcurveto{\pgfqpoint{6.356312in}{0.466613in}}{\pgfqpoint{6.334090in}{0.466613in}}%
\pgfpathlineto{\pgfqpoint{0.122222in}{0.466613in}}%
\pgfpathquadraticcurveto{\pgfqpoint{0.100000in}{0.466613in}}{\pgfqpoint{0.100000in}{0.444390in}}%
\pgfpathlineto{\pgfqpoint{0.100000in}{0.122222in}}%
\pgfpathquadraticcurveto{\pgfqpoint{0.100000in}{0.100000in}}{\pgfqpoint{0.122222in}{0.100000in}}%
\pgfpathclose%
\pgfusepath{stroke,fill}%
\end{pgfscope}%
\begin{pgfscope}%
\pgfsetroundcap%
\pgfsetroundjoin%
\pgfsetlinewidth{1.003750pt}%
\definecolor{currentstroke}{rgb}{0.298039,0.447059,0.690196}%
\pgfsetstrokecolor{currentstroke}%
\pgfsetdash{}{0pt}%
\pgfpathmoveto{\pgfqpoint{0.144444in}{0.377744in}}%
\pgfpathlineto{\pgfqpoint{0.366667in}{0.377744in}}%
\pgfusepath{stroke}%
\end{pgfscope}%
\begin{pgfscope}%
\definecolor{textcolor}{rgb}{0.150000,0.150000,0.150000}%
\pgfsetstrokecolor{textcolor}%
\pgfsetfillcolor{textcolor}%
\pgftext[x=0.455556in,y=0.338855in,left,base]{\color{textcolor}\rmfamily\fontsize{8.000000}{9.600000}\selectfont Soft Medoid GDC (\(\displaystyle T=1.0\))}%
\end{pgfscope}%
\begin{pgfscope}%
\pgfsetroundcap%
\pgfsetroundjoin%
\pgfsetlinewidth{1.003750pt}%
\definecolor{currentstroke}{rgb}{0.866667,0.517647,0.321569}%
\pgfsetstrokecolor{currentstroke}%
\pgfsetdash{}{0pt}%
\pgfpathmoveto{\pgfqpoint{0.144444in}{0.211104in}}%
\pgfpathlineto{\pgfqpoint{0.366667in}{0.211104in}}%
\pgfusepath{stroke}%
\end{pgfscope}%
\begin{pgfscope}%
\definecolor{textcolor}{rgb}{0.150000,0.150000,0.150000}%
\pgfsetstrokecolor{textcolor}%
\pgfsetfillcolor{textcolor}%
\pgftext[x=0.455556in,y=0.172215in,left,base]{\color{textcolor}\rmfamily\fontsize{8.000000}{9.600000}\selectfont Soft Medoid GDC (\(\displaystyle T=0.2\))}%
\end{pgfscope}%
\begin{pgfscope}%
\pgfsetroundcap%
\pgfsetroundjoin%
\pgfsetlinewidth{1.003750pt}%
\definecolor{currentstroke}{rgb}{0.333333,0.658824,0.407843}%
\pgfsetstrokecolor{currentstroke}%
\pgfsetdash{}{0pt}%
\pgfpathmoveto{\pgfqpoint{2.138942in}{0.377744in}}%
\pgfpathlineto{\pgfqpoint{2.361165in}{0.377744in}}%
\pgfusepath{stroke}%
\end{pgfscope}%
\begin{pgfscope}%
\definecolor{textcolor}{rgb}{0.150000,0.150000,0.150000}%
\pgfsetstrokecolor{textcolor}%
\pgfsetfillcolor{textcolor}%
\pgftext[x=2.450054in,y=0.338855in,left,base]{\color{textcolor}\rmfamily\fontsize{8.000000}{9.600000}\selectfont Soft Medoid GCN (\(\displaystyle T=50\))}%
\end{pgfscope}%
\begin{pgfscope}%
\pgfsetbuttcap%
\pgfsetroundjoin%
\pgfsetlinewidth{1.003750pt}%
\definecolor{currentstroke}{rgb}{0.768627,0.305882,0.321569}%
\pgfsetstrokecolor{currentstroke}%
\pgfsetdash{{1.000000pt}{1.650000pt}}{0.000000pt}%
\pgfpathmoveto{\pgfqpoint{2.138942in}{0.216640in}}%
\pgfpathlineto{\pgfqpoint{2.361165in}{0.216640in}}%
\pgfusepath{stroke}%
\end{pgfscope}%
\begin{pgfscope}%
\definecolor{textcolor}{rgb}{0.150000,0.150000,0.150000}%
\pgfsetstrokecolor{textcolor}%
\pgfsetfillcolor{textcolor}%
\pgftext[x=2.450054in,y=0.177751in,left,base]{\color{textcolor}\rmfamily\fontsize{8.000000}{9.600000}\selectfont Vanilla GCN}%
\end{pgfscope}%
\begin{pgfscope}%
\pgfsetbuttcap%
\pgfsetroundjoin%
\pgfsetlinewidth{1.003750pt}%
\definecolor{currentstroke}{rgb}{0.505882,0.447059,0.701961}%
\pgfsetstrokecolor{currentstroke}%
\pgfsetdash{{1.000000pt}{1.650000pt}}{0.000000pt}%
\pgfpathmoveto{\pgfqpoint{4.098950in}{0.377744in}}%
\pgfpathlineto{\pgfqpoint{4.321172in}{0.377744in}}%
\pgfusepath{stroke}%
\end{pgfscope}%
\begin{pgfscope}%
\definecolor{textcolor}{rgb}{0.150000,0.150000,0.150000}%
\pgfsetstrokecolor{textcolor}%
\pgfsetfillcolor{textcolor}%
\pgftext[x=4.410061in,y=0.338855in,left,base]{\color{textcolor}\rmfamily\fontsize{8.000000}{9.600000}\selectfont Vanilla GDC}%
\end{pgfscope}%
\begin{pgfscope}%
\pgfsetbuttcap%
\pgfsetroundjoin%
\pgfsetlinewidth{1.003750pt}%
\definecolor{currentstroke}{rgb}{0.576471,0.470588,0.376471}%
\pgfsetstrokecolor{currentstroke}%
\pgfsetdash{{3.700000pt}{1.600000pt}}{0.000000pt}%
\pgfpathmoveto{\pgfqpoint{4.098950in}{0.222811in}}%
\pgfpathlineto{\pgfqpoint{4.321172in}{0.222811in}}%
\pgfusepath{stroke}%
\end{pgfscope}%
\begin{pgfscope}%
\definecolor{textcolor}{rgb}{0.150000,0.150000,0.150000}%
\pgfsetstrokecolor{textcolor}%
\pgfsetfillcolor{textcolor}%
\pgftext[x=4.410061in,y=0.183922in,left,base]{\color{textcolor}\rmfamily\fontsize{8.000000}{9.600000}\selectfont SVD GCN}%
\end{pgfscope}%
\begin{pgfscope}%
\pgfsetbuttcap%
\pgfsetroundjoin%
\pgfsetlinewidth{1.003750pt}%
\definecolor{currentstroke}{rgb}{0.854902,0.545098,0.764706}%
\pgfsetstrokecolor{currentstroke}%
\pgfsetdash{{3.700000pt}{1.600000pt}}{0.000000pt}%
\pgfpathmoveto{\pgfqpoint{5.300079in}{0.377744in}}%
\pgfpathlineto{\pgfqpoint{5.522301in}{0.377744in}}%
\pgfusepath{stroke}%
\end{pgfscope}%
\begin{pgfscope}%
\definecolor{textcolor}{rgb}{0.150000,0.150000,0.150000}%
\pgfsetstrokecolor{textcolor}%
\pgfsetfillcolor{textcolor}%
\pgftext[x=5.611190in,y=0.338855in,left,base]{\color{textcolor}\rmfamily\fontsize{8.000000}{9.600000}\selectfont Jaccard GCN}%
\end{pgfscope}%
\begin{pgfscope}%
\pgfsetbuttcap%
\pgfsetroundjoin%
\pgfsetlinewidth{1.003750pt}%
\definecolor{currentstroke}{rgb}{0.549020,0.549020,0.549020}%
\pgfsetstrokecolor{currentstroke}%
\pgfsetdash{{3.700000pt}{1.600000pt}}{0.000000pt}%
\pgfpathmoveto{\pgfqpoint{5.300079in}{0.222811in}}%
\pgfpathlineto{\pgfqpoint{5.522301in}{0.222811in}}%
\pgfusepath{stroke}%
\end{pgfscope}%
\begin{pgfscope}%
\definecolor{textcolor}{rgb}{0.150000,0.150000,0.150000}%
\pgfsetstrokecolor{textcolor}%
\pgfsetfillcolor{textcolor}%
\pgftext[x=5.611190in,y=0.183922in,left,base]{\color{textcolor}\rmfamily\fontsize{8.000000}{9.600000}\selectfont RGCN}%
\end{pgfscope}%
\end{pgfpicture}%
\makeatother%
\endgroup%
}
\\
\vspace{-10pt}
\centering
  \makebox[\columnwidth][c]{
    \(\begin{array}{cc} 
    \subfloat[]{\resizebox{0.384\linewidth}{!}{\input{assets/cert_ratio_gdc_wide_del_no_legend_1.pgf}}} &
    \subfloat[]{\resizebox{0.361\linewidth}{!}{\input{assets/cert_ratio_gdc_wide_add_no_leglab_1.pgf}}} \\
    \subfloat[]{\resizebox{0.384\linewidth}{!}{\input{assets/cert_ratio_gdc_wide_del_no_legend_2.pgf}}} &
    \subfloat[]{\resizebox{0.361\linewidth}{!}{\input{assets/cert_ratio_gdc_wide_add_no_leglab_3.pgf}}} \\
    \end{array}\)
  }
  \caption{(a) and (b) show the certification ratio over different radii for deletion \(r_d\) and addition \(r_a\), for a combined noise of \(p_{-}=0.4\) and  \(p_{+}=0.001\). (c) shows the case of only deleting edges (\(p_{-}=0.4\), \(p_{+}=0\)) and (d) only adding edges (\(p_{-}=0\), \(p_{+}=0.001\)). For each plot we set the contrary radius to zero (e.g.\ in (a) \(r_a=0\)). We compare our Soft Medoid GDC against a GCN and the other defenses~\citep{Entezari2020, Wu2019, Zhu2019}. All plots are for Cora ML.\label{fig:appendix_certratio}}
\end{figure}

\begin{figure}[H]
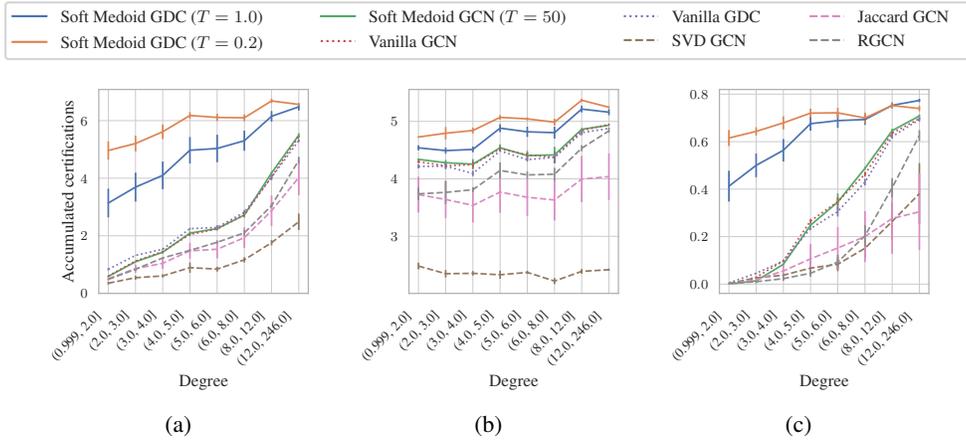

\centering
\resizebox{0.95\linewidth}{!}{
\begingroup%
\makeatletter%
\begin{pgfpicture}%
\pgfpathrectangle{\pgfpointorigin}{\pgfqpoint{6.456312in}{0.566613in}}%
\pgfusepath{use as bounding box, clip}%
\begin{pgfscope}%
\pgfsetbuttcap%
\pgfsetmiterjoin%
\definecolor{currentfill}{rgb}{1.000000,1.000000,1.000000}%
\pgfsetfillcolor{currentfill}%
\pgfsetlinewidth{0.000000pt}%
\definecolor{currentstroke}{rgb}{1.000000,1.000000,1.000000}%
\pgfsetstrokecolor{currentstroke}%
\pgfsetdash{}{0pt}%
\pgfpathmoveto{\pgfqpoint{0.000000in}{0.000000in}}%
\pgfpathlineto{\pgfqpoint{6.456312in}{0.000000in}}%
\pgfpathlineto{\pgfqpoint{6.456312in}{0.566612in}}%
\pgfpathlineto{\pgfqpoint{0.000000in}{0.566612in}}%
\pgfpathclose%
\pgfusepath{fill}%
\end{pgfscope}%
\begin{pgfscope}%
\pgfsetbuttcap%
\pgfsetmiterjoin%
\definecolor{currentfill}{rgb}{1.000000,1.000000,1.000000}%
\pgfsetfillcolor{currentfill}%
\pgfsetfillopacity{0.800000}%
\pgfsetlinewidth{1.003750pt}%
\definecolor{currentstroke}{rgb}{0.800000,0.800000,0.800000}%
\pgfsetstrokecolor{currentstroke}%
\pgfsetstrokeopacity{0.800000}%
\pgfsetdash{}{0pt}%
\pgfpathmoveto{\pgfqpoint{0.122222in}{0.100000in}}%
\pgfpathlineto{\pgfqpoint{6.334090in}{0.100000in}}%
\pgfpathquadraticcurveto{\pgfqpoint{6.356312in}{0.100000in}}{\pgfqpoint{6.356312in}{0.122222in}}%
\pgfpathlineto{\pgfqpoint{6.356312in}{0.444390in}}%
\pgfpathquadraticcurveto{\pgfqpoint{6.356312in}{0.466613in}}{\pgfqpoint{6.334090in}{0.466613in}}%
\pgfpathlineto{\pgfqpoint{0.122222in}{0.466613in}}%
\pgfpathquadraticcurveto{\pgfqpoint{0.100000in}{0.466613in}}{\pgfqpoint{0.100000in}{0.444390in}}%
\pgfpathlineto{\pgfqpoint{0.100000in}{0.122222in}}%
\pgfpathquadraticcurveto{\pgfqpoint{0.100000in}{0.100000in}}{\pgfqpoint{0.122222in}{0.100000in}}%
\pgfpathclose%
\pgfusepath{stroke,fill}%
\end{pgfscope}%
\begin{pgfscope}%
\pgfsetroundcap%
\pgfsetroundjoin%
\pgfsetlinewidth{1.003750pt}%
\definecolor{currentstroke}{rgb}{0.298039,0.447059,0.690196}%
\pgfsetstrokecolor{currentstroke}%
\pgfsetdash{}{0pt}%
\pgfpathmoveto{\pgfqpoint{0.144444in}{0.377744in}}%
\pgfpathlineto{\pgfqpoint{0.366667in}{0.377744in}}%
\pgfusepath{stroke}%
\end{pgfscope}%
\begin{pgfscope}%
\definecolor{textcolor}{rgb}{0.150000,0.150000,0.150000}%
\pgfsetstrokecolor{textcolor}%
\pgfsetfillcolor{textcolor}%
\pgftext[x=0.455556in,y=0.338855in,left,base]{\color{textcolor}\rmfamily\fontsize{8.000000}{9.600000}\selectfont Soft Medoid GDC (\(\displaystyle T=1.0\))}%
\end{pgfscope}%
\begin{pgfscope}%
\pgfsetroundcap%
\pgfsetroundjoin%
\pgfsetlinewidth{1.003750pt}%
\definecolor{currentstroke}{rgb}{0.866667,0.517647,0.321569}%
\pgfsetstrokecolor{currentstroke}%
\pgfsetdash{}{0pt}%
\pgfpathmoveto{\pgfqpoint{0.144444in}{0.211104in}}%
\pgfpathlineto{\pgfqpoint{0.366667in}{0.211104in}}%
\pgfusepath{stroke}%
\end{pgfscope}%
\begin{pgfscope}%
\definecolor{textcolor}{rgb}{0.150000,0.150000,0.150000}%
\pgfsetstrokecolor{textcolor}%
\pgfsetfillcolor{textcolor}%
\pgftext[x=0.455556in,y=0.172215in,left,base]{\color{textcolor}\rmfamily\fontsize{8.000000}{9.600000}\selectfont Soft Medoid GDC (\(\displaystyle T=0.2\))}%
\end{pgfscope}%
\begin{pgfscope}%
\pgfsetroundcap%
\pgfsetroundjoin%
\pgfsetlinewidth{1.003750pt}%
\definecolor{currentstroke}{rgb}{0.333333,0.658824,0.407843}%
\pgfsetstrokecolor{currentstroke}%
\pgfsetdash{}{0pt}%
\pgfpathmoveto{\pgfqpoint{2.138942in}{0.377744in}}%
\pgfpathlineto{\pgfqpoint{2.361165in}{0.377744in}}%
\pgfusepath{stroke}%
\end{pgfscope}%
\begin{pgfscope}%
\definecolor{textcolor}{rgb}{0.150000,0.150000,0.150000}%
\pgfsetstrokecolor{textcolor}%
\pgfsetfillcolor{textcolor}%
\pgftext[x=2.450054in,y=0.338855in,left,base]{\color{textcolor}\rmfamily\fontsize{8.000000}{9.600000}\selectfont Soft Medoid GCN (\(\displaystyle T=50\))}%
\end{pgfscope}%
\begin{pgfscope}%
\pgfsetbuttcap%
\pgfsetroundjoin%
\pgfsetlinewidth{1.003750pt}%
\definecolor{currentstroke}{rgb}{0.768627,0.305882,0.321569}%
\pgfsetstrokecolor{currentstroke}%
\pgfsetdash{{1.000000pt}{1.650000pt}}{0.000000pt}%
\pgfpathmoveto{\pgfqpoint{2.138942in}{0.216640in}}%
\pgfpathlineto{\pgfqpoint{2.361165in}{0.216640in}}%
\pgfusepath{stroke}%
\end{pgfscope}%
\begin{pgfscope}%
\definecolor{textcolor}{rgb}{0.150000,0.150000,0.150000}%
\pgfsetstrokecolor{textcolor}%
\pgfsetfillcolor{textcolor}%
\pgftext[x=2.450054in,y=0.177751in,left,base]{\color{textcolor}\rmfamily\fontsize{8.000000}{9.600000}\selectfont Vanilla GCN}%
\end{pgfscope}%
\begin{pgfscope}%
\pgfsetbuttcap%
\pgfsetroundjoin%
\pgfsetlinewidth{1.003750pt}%
\definecolor{currentstroke}{rgb}{0.505882,0.447059,0.701961}%
\pgfsetstrokecolor{currentstroke}%
\pgfsetdash{{1.000000pt}{1.650000pt}}{0.000000pt}%
\pgfpathmoveto{\pgfqpoint{4.098950in}{0.377744in}}%
\pgfpathlineto{\pgfqpoint{4.321172in}{0.377744in}}%
\pgfusepath{stroke}%
\end{pgfscope}%
\begin{pgfscope}%
\definecolor{textcolor}{rgb}{0.150000,0.150000,0.150000}%
\pgfsetstrokecolor{textcolor}%
\pgfsetfillcolor{textcolor}%
\pgftext[x=4.410061in,y=0.338855in,left,base]{\color{textcolor}\rmfamily\fontsize{8.000000}{9.600000}\selectfont Vanilla GDC}%
\end{pgfscope}%
\begin{pgfscope}%
\pgfsetbuttcap%
\pgfsetroundjoin%
\pgfsetlinewidth{1.003750pt}%
\definecolor{currentstroke}{rgb}{0.576471,0.470588,0.376471}%
\pgfsetstrokecolor{currentstroke}%
\pgfsetdash{{3.700000pt}{1.600000pt}}{0.000000pt}%
\pgfpathmoveto{\pgfqpoint{4.098950in}{0.222811in}}%
\pgfpathlineto{\pgfqpoint{4.321172in}{0.222811in}}%
\pgfusepath{stroke}%
\end{pgfscope}%
\begin{pgfscope}%
\definecolor{textcolor}{rgb}{0.150000,0.150000,0.150000}%
\pgfsetstrokecolor{textcolor}%
\pgfsetfillcolor{textcolor}%
\pgftext[x=4.410061in,y=0.183922in,left,base]{\color{textcolor}\rmfamily\fontsize{8.000000}{9.600000}\selectfont SVD GCN}%
\end{pgfscope}%
\begin{pgfscope}%
\pgfsetbuttcap%
\pgfsetroundjoin%
\pgfsetlinewidth{1.003750pt}%
\definecolor{currentstroke}{rgb}{0.854902,0.545098,0.764706}%
\pgfsetstrokecolor{currentstroke}%
\pgfsetdash{{3.700000pt}{1.600000pt}}{0.000000pt}%
\pgfpathmoveto{\pgfqpoint{5.300079in}{0.377744in}}%
\pgfpathlineto{\pgfqpoint{5.522301in}{0.377744in}}%
\pgfusepath{stroke}%
\end{pgfscope}%
\begin{pgfscope}%
\definecolor{textcolor}{rgb}{0.150000,0.150000,0.150000}%
\pgfsetstrokecolor{textcolor}%
\pgfsetfillcolor{textcolor}%
\pgftext[x=5.611190in,y=0.338855in,left,base]{\color{textcolor}\rmfamily\fontsize{8.000000}{9.600000}\selectfont Jaccard GCN}%
\end{pgfscope}%
\begin{pgfscope}%
\pgfsetbuttcap%
\pgfsetroundjoin%
\pgfsetlinewidth{1.003750pt}%
\definecolor{currentstroke}{rgb}{0.549020,0.549020,0.549020}%
\pgfsetstrokecolor{currentstroke}%
\pgfsetdash{{3.700000pt}{1.600000pt}}{0.000000pt}%
\pgfpathmoveto{\pgfqpoint{5.300079in}{0.222811in}}%
\pgfpathlineto{\pgfqpoint{5.522301in}{0.222811in}}%
\pgfusepath{stroke}%
\end{pgfscope}%
\begin{pgfscope}%
\definecolor{textcolor}{rgb}{0.150000,0.150000,0.150000}%
\pgfsetstrokecolor{textcolor}%
\pgfsetfillcolor{textcolor}%
\pgftext[x=5.611190in,y=0.183922in,left,base]{\color{textcolor}\rmfamily\fontsize{8.000000}{9.600000}\selectfont RGCN}%
\end{pgfscope}%
\end{pgfpicture}%
\makeatother%
\endgroup%
}
\\
\vspace{-10pt}
  \centering
  \makebox[\columnwidth][c]{
    \(\begin{array}{ccc} 
    \subfloat[]{\resizebox{0.27\linewidth}{!}{\input{assets/cert_area_over_deg_gdc_no_legend_1.pgf}}} &
    \subfloat[]{\resizebox{0.27\linewidth}{!}{\input{assets/cert_area_over_deg_gdc_no_leglab_2.pgf}}} &
    \subfloat[]{\resizebox{0.27\linewidth}{!}{\input{assets/cert_area_over_deg_gdc_no_leglab_3.pgf}}} \\
    \end{array}\)
  }
  \caption{(a) shows the accumulated certifications over the degree (equal frequency binning), for a combined noise of \(p_{-}=0.4\) and \(p_{+}=0.001\). (b) shows the case of only deleting edges (\(p_{-}=0.4\), \(p_{+}=0\)) and (c) only adding edges (\(p_{-}=0\), \(p_{+}=0.001\)). We compare our Soft Medoid GDC against a GCN and the other defenses~\citep{Entezari2020, Wu2019, Zhu2019}. All plots are for Cora ML.\label{fig:appendix_accumcertsoverdeg}}
\end{figure}

We complement the certification rates with the accumulated certifications for different node degrees in~\autoref{fig:appendix_accumcertsoverdeg} (compare to~\autoref{fig:accumcertsoverdegnogdc}). Especially in the case of only adding edges we see the strength of our approach. For all the other approaches~\cite{Kipf2017, Klicpera2019a, Entezari2020, Wu2019, Zhu2019} we can basically certify 0 \% of the low-degree nodes (degree \(\le\) 2 before adding self-loops). In contrast with the Soft Medoid GDC, we are able to certify around 50\% of the low-degree nodes!

In~\autoref{fig:appendix_certratio} we plot the certification ratios similarly to~\autoref{fig:certratio}. We can see that the Soft Medoid GDC outperforms the other approaches by a large margin. Especially figures (b) and (d) highlight the unparalleled difficulty of adversarially added edges. Also in the other cases \autoref{fig:appendix_certratio}(c) only deleting and (d) only adding edges, the Soft Medoid GDC clearly outperforms the other architectures. The margin is especially large in the challenging case (d) of solely adding edges.

\end{document}